\documentclass{article}

\usepackage{xcolor} 

\usepackage[final]{corl_2020} 

\usepackage{glossaries}
\usepackage{amssymb}
\usepackage{cleveref}
\usepackage{graphicx}
\usepackage{float}
\usepackage{caption}
\usepackage{subcaption}
\usepackage[ruled]{algorithm2e}
\usepackage{multirow}
\usepackage{tikz}
\usetikzlibrary{shapes,arrows,positioning,calc,patterns,fit}
\usepackage{etoolbox}
\usepackage[toc,page]{appendix}

\title{The Emergence of Adversarial Communication in Multi-Agent Reinforcement Learning}

\author{
  Jan Blumenkamp\\
  Department of Computer Science\\
  and Technology\\
  University of Cambridge\\
  United Kingdom\\
  \texttt{jb2270@cam.ac.uk} \\
  \And
  Amanda Prorok \\
  Department of Computer Science\\
  and Technology\\
  University of Cambridge\\
  United Kingdom\\
  \texttt{asp45@cam.ac.uk} \\
}

\input{symbols.sty}
\newacronym{AGNN}{AGNN}{Aggregation Graph Neural Network}
\newacronym{RL}{RL}{Reinforcement Learning}
\newacronym[longplural={Markov Decision Processes}]{MDP}{MDP}{Markov Decision Process}
\newacronym[longplural={Partially Observable Markov Decision Processes}]{POMDP}{POMDP}{Partially Observable Markov Decision Process}
\newacronym[longplural={Decentralized Partially Observable Markov Decision Processes}]{DecPOMDP}{Dec-POMDP}{Decentralized POMDP}
\newacronym{NN}{NN}{Neural Network}
\newacronym{CNN}{CNN}{Convolutional Neural Network}
\newacronym{GSO}{GSO}{Graph Shift Operator}
\newacronym{GPS}{GPS}{Global Positioning System}
\newacronym{GNN}{GNN}{Graph Neural Network}
\newacronym{PBT}{PBT}{Population Based Training}
\newacronym{PG}{PG}{Policy Gradient}
\newacronym{MLP}{MLP}{Multi Layer Perceptron}
\newacronym{LSTM}{LSTM}{Long Short-term Memory}
\newacronym{PPO}{PPO}{Proximal Policy Optimization}
\newacronym{mAP}{mAP}{mean Average Precision}
\newacronym{TRPO}{TRPO}{Trust Region Policy Optimization}
\newacronym{MARL}{MARL}{Multi-Agent Reinforcement Learning}
\newacronym{DQN}{DQN}{Deep Q-Learning}
\newacronym{GCNN}{GCNN}{Graph Convolutional Neural Network}
\newacronym{CPP}{CPP}{Coverage Path Planning}
\newacronym{VPG}{VPG}{Vanilla Policy Gradient}

\glsdisablehyper 

\makeatletter
\patchcmd\HyRef@autosetref{\bfseries ??}{\missingreftext}{}{\fail}
\newcommand\missingreftext{Supplemental Material}
\makeatother

\begin{document}
\maketitle

\begin{abstract}
Many real-world problems require the coordination of multiple autonomous agents. Recent work has shown the promise of Graph Neural Networks (GNNs) to learn explicit communication strategies that enable complex multi-agent coordination. These works use models of \textit{cooperative} multi-agent systems whereby agents strive to achieve a shared global goal. When considering agents with self-interested local objectives, the standard design choice is to model these as separate learning systems (albeit sharing the same environment). Such a design choice, however, precludes the existence of a single, differentiable communication channel, and consequently prohibits the learning of inter-agent communication strategies.
In this work, we address this gap by presenting a learning model that accommodates individual non-shared rewards and a differentiable communication channel that is common among all agents. We focus on the case where agents have self-interested objectives, and develop a learning algorithm that elicits the emergence of adversarial communications. We perform experiments on multi-agent coverage and path planning problems, and employ a post-hoc interpretability technique to visualize the messages that agents communicate to each other. We show how a single self-interested agent is capable of learning highly manipulative communication strategies that allows it to significantly outperform a cooperative team of agents.
\end{abstract}



\section{Introduction}

Multi-agent reinforcement learning models arise as a natural solution to problems where a common environment is influenced by the joint actions of multiple decision-making agents. Such solutions have been applied to a number of domains including traffic systems~\cite{wu_2017,hyldmar_2019}, dynamic supply-demand matching~\cite{nguyen_2018}, and multi-robot control~\cite{khan_2020, paulos_2019}.
In fully decentralized systems, agents not only need to learn how to behave cooperatively, but also, how to \textit{communicate} to most effectively coordinate their actions in pursuit of a common goal. 
Even though effective communication is key to successful decentralized coordination, the way a problem benefits from communication is not necessarily predetermined, especially in complex multi-agent settings where the optimal strategy is unknown.
Reinforcement learning has become one of the most promising avenues to solve such problems.

Capturing information that enables complex inter-agent coordination requires new kinds of \Gls{NN} architectures. \Glspl{GNN} exploit the fact that inter-agent relationships can be represented as graphs, which provide a mathematical description of the network topology. 
In multi-agent systems, an agent is modeled as a node in the graph, the connectivity of agents as edges, and the internal state of an agent as a graph signal. In recent years, a range of approaches towards learning explicit communication were made~\cite{foerster_2016,li_2020,tolstaya_2020,prorok2018graph}. 
The key attribute of \Glspl{GNN} is that they operate in a localized manner, whereby information is shared over a multi-hop communication network through explicit communication with nearby neighbors only, hence resulting in fully decentralizable policies.
 
One particularly promising approach leverages \Glspl{GCNN}, which utilize \textit{graph convolutions} to incorporate a graph structure into the learning process by concatenating layers of graph convolutions and nonlinearities~\cite{gama_2020, khan_2020}.
Recent work leverages \Glspl{GCNN} to automatically synthesize local communication and decision-making policies for solving complex multi-agent coordination problems~\cite{li_2020, tolstaya_2020}. These learning approaches assume full cooperation, whereby all agents share the same goal of maximising a global reward. 
Yet there is a dearth of work that explores whether agents can utilize machine learning to synthesize communication policies that are not only cooperative, but instead, are \textit{non-cooperative} or even \textit{adversarial}. The goal of this paper is to demonstrate how self-interested agents can learn such adversarial communication, without explicitly optimizing for it.
Crucially, we posit that understanding how adversarial communication emerges is the first step towards developing methods that can deal with it in real-world situations.

\textbf{Contributions.}
The main contribution of this paper is an evaluation of the hypothesis that non-cooperative agents can learn manipulative communication policies. To achieve this goal, we developed a new multi-agent learning model that integrates heterogeneous, potentially self-interested policies that share a differentiable communication channel. 
This model consists of three key components, \textit{(i)} a monolithic, decentralizable neural architecture that accommodates multiple distinct reward functions and a common differentiable communication channel, \textit{(ii)} a reinforcement learning algorithm that elicits the emergence of adversarial communications, and \textit{(iii)}, a post-hoc interpretability technique that enables the visualization of communicated messages.
Our code is publicly available~\footnote{\url{https://github.com/proroklab/adversarial_comms}}.

The experimental evaluation is based on a multi-agent system with a mix of cooperative and self-interested agents, and demonstrates the effectiveness of the learning scheme in
multi-agent coverage and path planning problems.
Results show that it is possible to learn highly effective communication strategies capable of manipulating other agents to behave in such a way that it benefits the self-interested agents.
Overall, we demonstrate that adversarial communication emerges when local rewards are drawn from a finite pool, or when resources are in contention. We also show that self-interested agents that communicate manipulatively, however, need not be adversarial by design; they are simply programmed to disregard other agents' rewards.

\section{Related Work}

We briefly review work in cooperative and non-cooperative multi-agent reinforcement learning, with a focus on approaches that model communication between agents. 

\textbf{\textit{Cooperative} multi-agent reinforcement learning}.
Cooperation enables agents to achieve feats together that no individual agent can achieve on its own.
Yet independently learning agents perform poorly in practice~\cite{matignon_2012}, since agents' policies change during training, resulting in a non-stationary environment. Hence, the majority of recent work leans on joint learning paradigms~\cite{gupta_2017, rashid_2018, foerster_2018, omidshafiei_2017}. These approaches avoid the need for explicit communication by making strong assumptions about the visibility of other agents and the environment. Some other approaches use communication, but with a predetermined protocol \cite{maravall_2011, zhang_2013}.

Early work on learning communication considers \textit{discrete} communication, through signal binarization~\cite{foerster_2016}, or categorical communication emissions~\cite{mordatch_2017}. The former approaches demonstrate emergent communication among few agents; scaling the learning process to larger agent teams requires innovations in the structure of the learnt communication models. 
The approach in~\cite{sukhbaatar_2016} presents a more scalable approach by instantiating a \Gls{GNN}-inspired construction for learning continuous communication.
Other, more recent work demonstrates the use of \Glspl{GNN} for learning communication policies that lead to successful multi-agent coordination in partially observed environments~\cite{tolstaya_2020, li_2020, khan_2020}.

\textbf{\textit{Non-cooperative} multi-agent reinforcement learning}.
Most work on non-cooperative multi-agent systems does not model \textit{learnable} communication policies, since the assumption is made that agent behaviors evolve as a function of consequences observed in the environment. \textit{Social dilemma} problems represent one type of non-cooperative system, where the collectively best outcomes are not aligned with individualistic decisions. 
Descriptive results were obtained for sequential social dilemmas~\cite{leibo_2017} and common-pool resource problems~\cite{perolat_2017}. Other work focuses on the development of learning algorithms for non-cooperative multi-player games~\cite{serrino_2019, paquette_2019}.
Yet none of these approaches include dedicated communication channels between agents.

More closely related to our work, the work in~\cite{lowe_2017} presents a learning scheme for mixed cooperative-competitive settings. The approach enables speaker agents to output semantic information, which is, in turn, observed by listener agents.
In contrast to our approach, this type of communication is not differentiable, and assumes time-invariant fully connected agent topologies. To date, there is a lack of work in non-cooperative multi-agent reinforcement learning with continuous differentiable communication.


\section{Preliminaries}

This work considers the presence of an explicit communication channel between agents. As such, we first introduce \Glspl{AGNN}~\cite{gama_2020}, which provide a decentralizable architecture to learn communication policies which are fully differentiable.~\footnote{An \Gls{AGNN} is a particular instantiation of \Glspl{GNN}~\cite{battaglia2016interaction,gilmer2017neural}. Our method can be used with any \Gls{GNN} variant.}
We then proceed with the introduction of \Gls{VPG}~\cite{schulman_2016} for groups of independent \emph{self-interested} agents.

\subsection{Aggregation Graph Neural Networks for Multi-Agent Communication}
\label{sec:gnn}

We model inter-agent communication through a graph $\ccalG=\langle \ccalV, \ccalE \rangle$.
The node set $\ccalV=\{1, \ldots, N\}$ represents individual agents, and the edge set $\ccalE=\ccalV \times \ccalV$ represents inter-agent communication links.
The set of neighboring agents $\ccalN_i$ that can communicate with agent $i \in \ccalV$ is defined as $\ccalN_i = \{j \in \ccalV: (j, i) \in \ccalE\}$.
The adjacency matrix $\bbS \in \reals^{N \times N}$ indicates the connectivity of the graph with the entry $[\bbS]_{ij}$ equal to one if node $j \in \ccalN_i$ (and zero otherwise).
We can make the dependency on the edge set $\ccalE$ explicit with $\bbS_\ccalE$.

The set of messages transmitted by all robots is denoted $\bbX \in \reals^{N \times F}$. Hence, the message (or \emph{datum}) sent by agent $i$ is $\bbx_i = [\bbX]_i$ (the $i$th row of matrix $\bbX$).
\Glspl{AGNN} operate over multiple communication hops $k \in {0, \ldots, K}$ and the connectivity at each hop $k$ can be computed by elevating the adjacency matrix to the power $k$ (i.e., $\bbS^k$).
For each hop $k$, messages are aggregated using a permutation-invariant operation (e.g., sum or average) and the next message is computed from this aggregated data.
More formally, the \emph{graph shift operator} $\bbS$ shifts the signal $\bbX$ over the nodes and a series of learnable \emph{filter taps} $\bbH_k \in \reals^{F \times F'}$ aggregates data from multiple hops, such that
\begin{align}
    [\bbS\bbX]_i = \sum_{j \in \ccalN_i} [\bbS]_{ij} \bbx_j  ~~~~\mathrm{and}~~~~  
    \bbX' = g_\eta(\bbX; \bbS) = \sum_{k=0}^K \bbS^k\bbX\bbH_k
    \label{eq:gnn_graph_conv}
\end{align}

where $g$ is a function parameterized by $\eta = \{ \bbH_k \}_{k=1}^K$ and where $\bbX'$ summarizes the data received by all agents.
A non-linearity $\sigma$ is applied and the process is cascaded $L$ times:
\begin{align}
    \bbX_l = \sigma \left( g_{\eta_l}(\bbX_{l-1}; \bbS) \right) \textrm{~with~} \bbX_0 = \bbX.
\end{align}
As indicated by the first equality in \autoref{eq:gnn_graph_conv}, this sequence of operations is decentralizable, and can be executed locally at each agent~\cite{gama_2020}.
We also note that, since $\eta_l$ for $l \in \{1, \ldots, L\}$ is shared among agents, this formulation can only accommodate \textit{homogeneous} teams of agents--- a limitation that we lift in~\autoref{sec:hetero_gnn} .


\subsection{Independent Vanilla Policy Gradient}
\label{sec:vpg}

We consider multi-agent systems that are partially observable and operate in a decentralized manner.
Each agent aims to maximise a local reward with no access to the true global state.
In \autoref{sec:coop}, we allow agents to communicate through the exchange of explicit messages.

\paragraph{\Gls{MDP}.}
We formulate a stochastic game defined by the tuple $\langle \ccalV, \ccalS, \ccalA, P, \{R^i\}_{i\in\ccalV}, \{\ccalZ^i\}_{i\in\ccalV}, Z, \ccalT, T, \gamma \rangle$, in which $N$ agents identified by $i \in \ccalV \equiv \{ 1, \ldots, N \}$ choose sequential actions.
Similarly to the formulation in~\cite{foerster_2018}, the environment has a true state $s_t \in \ccalS$.
At each time step $t$, each agent executes action $a^i_t \in \ccalA$. Together, the agents form a joint action $\bba_t \in \ccalA^N$.
The state transition probability function is $P(s_{t+1}|s_t, \bba_t)~:~\ccalS \times \ccalA^N \times \ccalS \rightarrow [0, 1]$.
We consider a partially observable setting, in which each agent $i$ draws observations $z_t^i \in \ccalZ^i$ according to the observation function $Z^i(s_t)~:~\ccalS \rightarrow \ccalZ$, forming a joint observation $\bbz_t \in \ccalZ^N$.
The discount factor is denoted by $\gamma \in [0, 1)$.
To the contrary of~\cite{foerster_2018}, agents in our system observe a local reward $r^i_t$ drawn from $R^i(s_t, a^i_t)~:~\ccalS \times \ccalA \rightarrow \reals$ (i.e., it is not a global reward given to the whole team). The agents' communication topology at time step $t$, denoted by $\ccalE_t \in \ccalT$, is drawn according to $T(s_t)~:~\ccalS \rightarrow \ccalT$. 

\paragraph{Multi-Agent Vanilla Policy Gradient (VPG).}
Next, we detail \Gls{VPG} applied to independent self-interested agents with a centralized critic---a setting similar to~\cite{foerster_2018}, but with individual agent-specific rewards.
Without explicit communication, each agent learns a local policy $\pi^i_{\theta^i}(a^i_t | z^i_t)~: \ccalZ \times \ccalA \rightarrow [0, 1]$ parameterized by a local set of parameters $\theta^i$.
The induced joint policy is $\bbpi_\theta(\bba_t | \bbz_t) = \prod_{i\in\ccalV} \pi^i_{\theta^i}(a^i_t | z^i_t)$ with $\theta = \{\theta^i\}_{i\in\ccalV}$.
The discounted return for each agent $i$ is $G^i_t = \sum_{l=0}^\infty \gamma^l r^i_{t+l}$.
The centralized value functions that estimate the value of each agent $i$ in state $s_t$ are $V^{\bbpi,i}(s_t) = \E_t[G^i_t | s_t]$, and the corresponding action-value functions are $Q^{\bbpi,i}(s_t, \bba_t) = \E_t[G^i_t | s_t, \bba_t]$ (we now omit $\bbpi$ and simply write $V^i$ and $Q^i$).
The advantage functions are then given by $A^i(s_t, \bba_t) = Q^i(s_t, \bba_t) - V^i(s_t)$.
There exist many techniques to estimate the advantage.
For example,~\citet{schulman_2016} define the \emph{Generalized Advantage Estimate} as $\hat{A}^i(s_t) = \sum_{l=0}^\infty (\gamma\lambda)^l \delta^i_{t+l}$ where $\lambda \in [0, 1]$ and where the TD-residual is $\delta^i_t = r^i_t + \gamma V^i(s_{t+1}) - V^i(s_t)$.
The value functions themselves can be estimated by parameterized functions $V_\phi^i$ (with parameters $\phi$) by minimizing $\E_t [\| V_\phi^i(s_t) - G^i_t \|^2]$ (again, there are many approaches to estimate value functions~\cite{bertsekas1995dynamic}).
However, since this is not the focus of this work, we will now assume that we have access to a centralized advantage estimate $\hat{A}^i(s_t)$ for each agent $i$.
Finally, the policy of agent $i$ can be improved through gradient ascent using the policy gradient
\begin{align}
    g^i = \E_t \left[ \nabla_\theta \log \pi^i_{\theta^i}(a_t^i | z^i_t) \hat{A}^i(s_t) \right].
\end{align}

\section{Methodology}
\definecolor{decentr_col}{rgb}{0.2,0.8,0.4}
\definecolor{centr_col}{rgb}{1,0.6,0}
\definecolor{coop_col}{rgb}{0,0,1}
\definecolor{adv_col}{rgb}{1,0,0}

\pgfdeclarelayer{bg}    
\pgfsetlayers{bg,main}  

\tikzset{
    centr/.style = {draw=centr_col},
    decentr/.style = {draw=decentr_col},
    cooperative/.style = {draw=coop_col, thick},
    adversarial/.style = {draw=adv_col, thick},
    block/.style = {draw, rectangle},
    tmp/.style  = {coordinate}, 
    concat/.style = {draw, circle, inner sep=0pt, minimum size=0.05cm},
    input/.style = {},
    output/.style = {},
    cnn/.style = {trapezium, trapezium angle=75, minimum width=0.5cm, draw},
    actor/.style = {draw},
    critic/.style = {draw},
    agent/.style 2 args={
        draw, dashed, inner sep=1.5em, rounded corners=0.2cm,
        label={[anchor=south east]south east:{$\pi^{#1}_{\theta}$}}
    },
    env/.style = {draw, rounded corners=0.1cm},
    alg/.style = {draw, rounded corners=0.1cm},
    comm/.style = {draw, fill=white, fill opacity=0.7, text opacity=1.0, dashed, rounded corners=0.1cm},
}

\newcommand{\architectureactor}{
    \begin{tikzpicture}[auto, decentr, node distance=1cm,>=latex']
        \node [tmp] (z_input) {};
        \node [cnn, above=0.5cm of z_input] (cnn) {CNN};
        \node [block, above of=cnn] (gnn) {GNN};
        \node [tmp, right=0.25cm of gnn] (m_gnn) {};
        \node [block, above of=gnn] (mlp) {MLP};
        \node [output, above=0.5cm of mlp] (a_output) {Actor $i$};
        

        \draw[->](z_input) -- node [pos=0.5, left]{$z_t^i$} (cnn);
        \draw[->](cnn) -- node {}(gnn);
        \draw[->](gnn) -- node {}(mlp);
        \draw[->](mlp) -- node [pos=0.5, left] {$a_t^i$}(a_output);
        \draw[->, transform canvas={yshift=0.5ex}](gnn) -- node {}(m_gnn);
        \draw[->, transform canvas={yshift=-0.5ex}](m_gnn) -- node {}(gnn);
    \end{tikzpicture}
}

\newcommand{\architecturecritic}{
    \begin{tikzpicture}[auto, centr, node distance=1cm,>=latex']
        \node [cnn] (cnn_z) {CNN};
        \node [tmp, name=z_input, below=0.5cm of cnn_z] (z_input) {};
        \node [cnn, right=0.25cm of cnn_z] (cnn_s) {CNN};
        \node [tmp, name=s_input, below=0.5cm of cnn_s] (s_input) {};
        \node [concat, above of=cnn_s] (concat) {$+$};
        \node [block, above of=concat] (mlp) {MLP};
        \node [output] (a_output) [above=0.5cm of mlp] {Critic $i$};
        
        \draw[->](s_input) -- node [pos=0.5, left] {$s_t$}(cnn_s);
        \draw[->](z_input) -- node [pos=0.5, left] {$z_t^i$}(cnn_z);
        \draw[->](cnn_s) -- node {}(concat);
        \draw[->](concat) -- node {}(mlp);
        \draw[->](cnn_z) |- node {}(concat);
        \draw[->](mlp) -- node [pos=0.5, left] {$V_t^i$}(a_output);
    \end{tikzpicture}
}

\newcommand{\architectureoverview}{
    \begin{tikzpicture}[auto, node distance=1cm,>=latex']
        \node [actor, decentr] (actor_coop) {Actor};
        \node [tmp, below right=0.25cm and 0.8em of actor_coop.south] (link_coop_actor_critic) {};
        \node [critic, centr, right=0.3cm of actor_coop] (critic_coop) {Critic};
        \node [agent={i}, label={[anchor=north west]north west:$i \in \ccalV$},
        label={[anchor=north]north:$i \neq n$},
        cooperative, fit={(actor_coop) (critic_coop)}] (all_coop) {};
        
        \node [actor, decentr, right=1.5cm of critic_coop] (actor_adv) {Actor};
        \node [tmp, below right=0.25cm and 0.8em of actor_adv.south] (link_adv_actor_critic) {};
        \node [critic, centr, right=0.3cm of actor_adv] (critic_adv) {Critic};
        \node [agent={n}, adversarial, fit={(actor_adv) (critic_adv)}] (all_adv) {};
        
        \path let
            \p1=(all_coop.west),
            \p2=(all_adv.east)
        in node [
            env,
            below=0.4cm of all_coop.south west,
            anchor=north west,
            minimum width=\x2-\x1-\pgflinewidth
        ] (env) {Environment};
        
        \path let
            \p1=(actor_coop.west),
            \p2=(actor_adv.east)
        in node [
            comm,
            decentr,
            above=1.25cm of actor_coop.north west,
            anchor=north west,
            minimum width=\x2-\x1-\pgflinewidth,
        ] (comm) {$\ccalE_t$ Communication Channel};
        
        \path let
            \p1=(critic_coop.west),
            \p2=(critic_adv.east)
        in node [
            alg,
            centr,
            above=2cm of critic_coop.north west,
            anchor=north west,
            minimum width=\x2-\x1-\pgflinewidth
        ] (alg) {Algorithm};

        
        \draw [->, decentr, transform canvas={xshift=0.5ex}] (actor_coop) -- (actor_coop |- comm.south);
        \draw [<-, decentr, transform canvas={xshift=-0.5ex}] (actor_coop) -- (actor_coop |- comm.south);
        
        \draw [->, decentr, transform canvas={xshift=-0.8em}] (actor_coop) --
            node[pos=0.25, left]{$a_t^i$}
            (actor_coop |- env.north);
        \draw [<-, decentr, transform canvas={xshift=0.8em}] (actor_coop) --
            node[pos=0.25, left]{$z_t^i$}
            (actor_coop |- env.north);

        \draw [->, centr] (link_coop_actor_critic) -| ([xshift=-0.5ex] critic_coop.south);
        \draw [<-, centr, transform canvas={xshift=0.5ex}] (critic_coop) -- node[pos=0.8, right]{$s_t$} (critic_coop |- env.north);
        
        \begin{pgfonlayer}{bg}
            \draw [->, centr] (critic_coop) -- node[pos=0.15, right]{$V_t^i$} (critic_coop |- alg.south);
        \end{pgfonlayer}

        \draw [->, decentr, transform canvas={xshift=0.5ex}] (actor_adv) -- (actor_adv |- comm.south);
        \draw [<-, decentr, transform canvas={xshift=-0.5ex}] (actor_adv) -- (actor_adv |- comm.south);

        \draw [->, decentr, transform canvas={xshift=-0.8em}] (actor_adv) --
            node[pos=0.25, left]{$a_t^n$}
            (actor_adv |- env.north);
        \draw [<-, decentr, transform canvas={xshift=0.8em}] (actor_adv) --
            node[pos=0.25, left]{$z_t^n$}
            (actor_adv |- env.north);
        
        \draw [->, centr] (link_adv_actor_critic) -| ([xshift=-0.5ex] critic_adv.south);
        \draw [<-, centr, transform canvas={xshift=0.5ex}] (critic_adv) -- node[pos=0.8, right]{$s_t$} (critic_adv |- env.north);
        
        \begin{pgfonlayer}{bg}
            \draw [->, centr] (env) -- node[pos=0.05, left]{$r_t$} (env |- alg.south);
        \end{pgfonlayer}

        \draw [->, centr] (critic_adv) --
            node[pos=0.15, right]{$V_t^n$}
            (critic_adv |- alg.south);

    \end{tikzpicture}
}

\begin{figure}[tb]
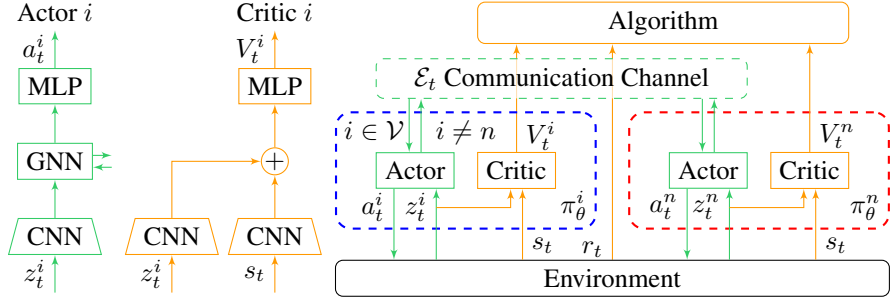

    \centering
    \architectureactor
    \architecturecritic
    \architectureoverview
    \caption{Architectural overview: The actor or the policy $\pi_\theta^i$ determines the action $a_t^i$ from the local observation $z_t^i$ and the multi-hop communication messages from other agents, which in turn depend on all agents' observations $z_t$ and the communication topology $\ccalE_t$. 
    Green denotes components that can be executed locally, and orange denotes components that are required for centralized training only. In this figure and the rest of this paper we highlight components and results for the cooperative team in blue and the self-interested agent in red.}
    \label{fig:architecture_overview}
\end{figure}

The objective of this work is to demonstrate the emergence of adversarial communication.
Towards this end, we consider a mixed setup with a team of cooperative agents and one self-interested agent. We first propose a modification to the \Gls{AGNN} elaborated in \autoref{sec:gnn}, to allow for multi-agent systems with heterogeneous policies.
We then modify \Gls{VPG} to train the team of cooperative agents to collaborate using explicit communication.
Finally, we introduce the self-interested agent and train it to communicate with the cooperative team.
An overview of the architecture is illustrated in~\autoref{fig:architecture_overview}; details of the algorithms are given in \autoref{sec:sup:learning_algorithms}.

\subsection{Heterogeneous AGNN}
\label{sec:hetero_gnn}

In order to use the \Gls{AGNN} architecture in a heterogeneous setting, we need to generalize the homogeneous formalization defined in \autoref{eq:gnn_graph_conv}.
In the homogeneous setting, a single set of parameters is given as $\eta = \{ \bbH_k \}_{k=1}^K$ where $\bbH_k$ describes a trainable \emph{filter tap} for hop $k$.
To allow for locally unique communication policies, we introduce different \emph{filter taps} for individual agents.
\autoref{eq:gnn_graph_conv} becomes
\begin{align}
    \bbX' = g_\eta(\bbX; \bbS) = \sum_{k=0}^K
    \begin{bmatrix}
        [\bbS^k]_1^\mathsf{T}\bbX\bbH^1_k \\
        \vdots \\
        [\bbS^k]_N^\mathsf{T}\bbX\bbH^N_k
    \end{bmatrix}
    \label{eq:gnn_multi_graph_conv}
\end{align}
where $\eta = \{ \eta^i \}_{i \in \ccalV}$ and $\eta^i = \{ \bbH^i_k \}_{k=1}^K$.
It is important to observe that each aggregation $[\bbS^k]_i^\mathsf{T}\bbX$ only considers messages sent by agents $k$ hops away from $i$ and that this formulation remains decentralizable.
Note that not all $\eta^i$ need to be distinct, i.e., a sub-group of agents can share the same communication policy.

\subsection{Cooperative Learning}
\label{sec:coop}

We first consider a team of cooperative homogeneous agents with individual rewards and explicit communication, and formulate a local decentralized policy.

\paragraph{Local Policy.}
Each agent encodes its local observation $z^i_t$ using an encoder $f_{\nu}(z^i_t)~:~\ccalZ \rightarrow \reals^F$ (e.g., using a \Gls{CNN}).
The encoded observations are grouped as $\bbX = [f_{\nu}(z^1_t), \ldots, f_{\nu}(z^N_t)]^\mathsf{T}$, and each local encoding is then shared with neighboring agents by applying $\bbX' = \sigma(g_\eta(\bbX, \bbS_{\ccalE_t}))~:~\reals^{N \times F} \times \reals^{N \times N} \rightarrow \reals^{N \times F'}$ (i.e., using an \Gls{AGNN}).
The aggregated data is then used to output a distribution over actions using $h_{\mu}([\bbX]_i)~:~\reals^{F'} \rightarrow \Delta^{|\ccalA|}$ (where $\Delta$ represents the simplex), e.g., using a \Gls{MLP} with a softmax output. 
Overall, the local policy is defined by $\pi_{\theta}^i(a_t^i | z_t, \ccalE_t) = h_{\mu}([\sigma(g_\eta([f_{\nu}(z^1_t), \ldots, f_{\nu}(z^N_t)])^\mathsf{T}, \bbS_{\ccalE_t})]_i)$ with $\theta = \{ \nu, \eta, \mu \}$ (we omit the explicit dependence on $\ccalE_t$ and write $\pi_{\theta}^i(a_t^i | z_t)$).
Despite being decentralized and locally executable, each agent's policy explicitly depends on the observations made by neighboring agents (due to the explicit communication). 

\paragraph{Cooperative Policy Gradient.}
To train the cooperative group of agents, we modify \gls{VPG} (\autoref{sec:vpg}) to reinforce local, individual actions that lead to increased rewards for other agents.
\begin{lemma}
Given an actor-critic algorithm with a compatible TD(1) critic that follows the cooperative policy gradient
\begin{align}
    g^i_k = \E_{\bbpi} \left[ \sum_{j\in\ccalV} \nabla_\theta \log \pi^j_{\theta}(a^j | z) A^i(s) \right]
\end{align}
for each agent $i \in \ccalV$ at each iteration $k$, this gradient converges to a local maximum of the expected sums of returns of all agents with probability one. 
\end{lemma}
The proof is provided in \autoref{sec:sup:proofs}. This lemma highlights that not only does the gradient converge, but also that the joint policy maximizes the sum of cumulative rewards.

\subsection{Self-Interested Learning}

After training the cooperative policy, we replace one of the agents with a self-interested agent.
This agent's goal is simply to maximize its own reward (disregarding the rewards of others). If rewards are drawn from a finite pool (e.g., agents compete for resources), we expect the self-interested agent to learn to communicate erroneous information. In other words, it will start lying about its state and observations to mislead other agents, thereby increasing its own access to the limited rewards.

\textbf{Local Policy.} For clarity, we denote the parameters of all cooperative agents by $\theta^c$ and the parameters of the self-interested agent by $\theta^n$.
The index of the self-interested agent is denoted by $n$.
The policy of each agent $i$ is denoted by $\pi_{\theta^c\theta^n}^i(a_t^i | z_t)$.
It depends on both $\theta^c$ and $\theta^n$ for all agents, as messages exchanged between agents are inter-dependent.

\textbf{Self-Interested Policy Gradient.} The goal of this learning procedure is to learn $\theta^n$ to maximize the expected return of the self-interested agent under policy $\pi_{\theta^c\theta^n}^n(a_t^n | z_t)$ where $\theta^c$ is fixed.
Similarly to \autoref{sec:coop}, we modify the policy gradient of the self-interested agent to account for its advantage across actions performed by other agents. 
\begin{lemma}
Given an actor-critic algorithm with a compatible TD(1) critic that follows the self-interested policy gradient
\begin{align}
    g^n_k = \E_{\bbpi} \left[ \sum_{j\in\ccalV} \nabla_{\theta^n} \log \pi^j_{\theta^c\theta^n}(a^j | z) A^n(s) \right]
\end{align}
for a self-interested agent $n \in \ccalV$ at each iteration $k$, this gradient converges to a local maximum of the expected returns of the self-interested agent with probability one. 
\end{lemma}
The proof is provided in \autoref{sec:sup:proofs}. Note how this gradient only affects the parameters $\theta^n$ of the self-interested policy.

\subsection{White-Box Analysis}
The encoder $f_{\nu^c}$ of the cooperative policy $\pi^i_{\theta^c}$, where $\nu^c$ refers to the parameters of the cooperative encoder, transforms the local observation $z_t^i$ into an encoding $[\bbX_t]_i$ which is used as input to the \Gls{AGNN} and constitutes the first message being sent by agent $i$ at time step $t$. 
We hypothesize that the self-interested agent creates an alternative encoding that not only helps it gather more rewards individually, but also, influences cooperative agents' behavior towards its self-interested goal.
To verify this hypothesis, we perform a white-box analysis by sampling observations $z_t^i$ and feature vectors $[\bbX_t]_i$ for all cooperative agents, and by training an interpreter $f^{-1}_\psi$ that minimizes the reconstruction error such that $f^{-1}_\psi \circ f_{\nu^c}(z_t^i) \approx z_t^i$ for all $z_t^i$. We provide more details in \autoref{sec:sup:impl_details}.


\section{Experiments}
\label{sec:experiments}

\begin{figure}[tb]
    \centering
    \begin{subfigure}[t]{0.32\textwidth}
        \centering
        \includegraphics[width=0.49\linewidth]{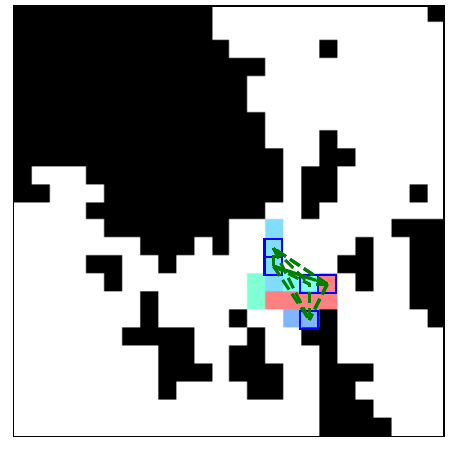}
        \includegraphics[width=0.49\linewidth]{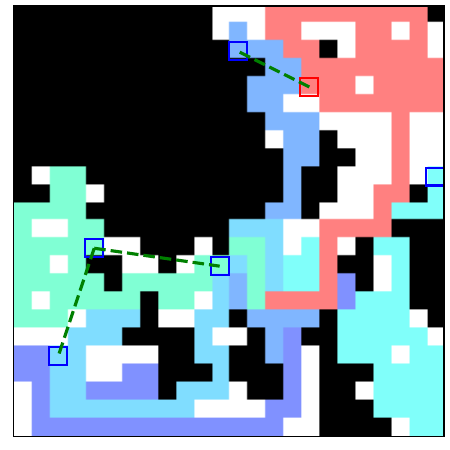}
        \caption{Standard coverage.}
        \label{fig:env_coverage}
    \end{subfigure}
    \hfill
    \begin{subfigure}[t]{0.32\textwidth}
        \centering
        \includegraphics[width=0.49\linewidth]{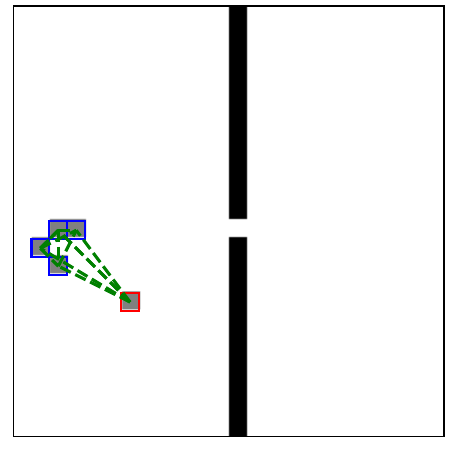}
        \includegraphics[width=0.49\linewidth]{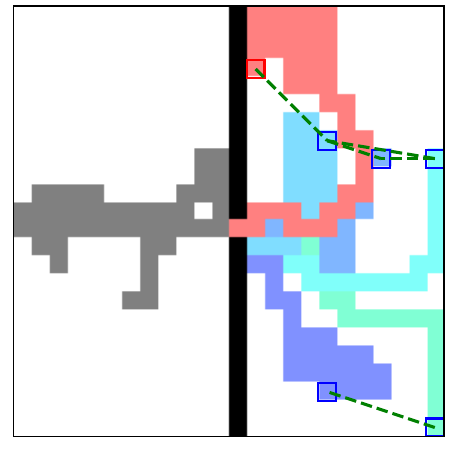}
        \caption{Coverage in split environment.}
        \label{fig:env_split}
    \end{subfigure}
    \hfill
    \begin{subfigure}[t]{0.32\textwidth}
    \centering
        \includegraphics[width=0.49\linewidth]{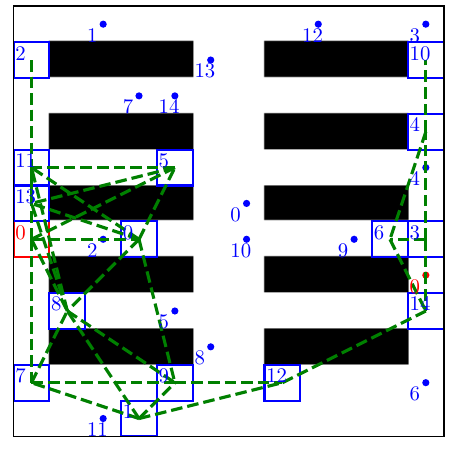}
        \includegraphics[width=0.49\linewidth]{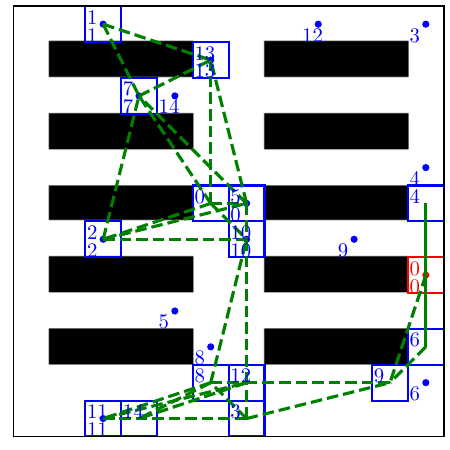}
        \caption{Path planning.}
        \label{fig:env_path}
    \end{subfigure}
    \caption{Overview of grid-world environments used in our experiments. Cooperative and self-interested agents are visualized as blue and red squares, respectively.
    Black cells correspond to obstacles. In the coverage environments, different colors indicate the coverage achieved by individual agents. In the path planning environment, labeled goal locations are indicated by circles.}
    \label{fig:environments}
\end{figure}

We validate our proposed learning scheme and architecture on three case studies requiring communication between agents. 
For all our experiments, we use a PPO variation of our algorithms~\cite{schulman_2017} and employ distributed training based on Ray~\cite{moritz_2018} and RLlib~\cite{liang_2018}. A movie of our experiments is available at \url{https://youtu.be/o1Nq9XoSU6U}.

\paragraph{Setup.}
We evaluate our learning scheme in a custom grid-world (see~\autoref{fig:environments}). Agents can communicate if they are closer than some predefined distance, thus defining the communication topology $\ccalE_t$.
Each agent has a local field of view. 
We perform three experiments, \textbf{(1)} with a purely cooperative team, \textbf{(2)} with the introduction of one self-interested agent holding the cooperative team's policy fixed, and \textbf{(3)}, with the cooperative team allowed to re-adapt to the self-interested agent. 
In experiment (2), we perform two variants, one where the self-interested agent can communicate to the other agents, and one where it cannot. We evaluate average agent performance, and use our white-box interpreter to understand the nature of messages sent.

\paragraph{Tasks.} We consider the following three tasks, as depicted in~\autoref{fig:environments}. More details on each task are given in \autoref{sec:sup:impl_details}.

\textit{Coverage in non-convex environments:} An agent is required to visit all free cells in the environment as quickly as possible; it is rewarded for moving into a cell that has not yet been covered by any other agent (including itself). The observation is described by a three-channel tensor consisting of the local obstacle map, the agent's own coverage, and the agent's position. We use $N=6$ agents.

\textit{Coverage in split environments:} An agent is required to visit all free cells in the \textit{right-hand} side of the environment as quickly as possible; it is rewarded for covering new cells in that sub-area. The observation is the same as for the prior task. We use $N=6$ agents.

\textit{Path planning:} An agent is required to navigate to its assigned goal. Only one agent can occupy any given cell at the same time, hence, agents must learn to avoid each other. An agent is rewarded for each time-step that it is located at its designated goal (episodes have fixed horizons). The observation is similar to the coverage tasks, except the channel containing the agent's coverage, which is replaced with a map containing its goal. We use $N=16$ agents.

\begin{table}[tb]
    \centering
    \small
    \begin{tabular}{l|c||c|c||c|c||c}
        \multicolumn{2}{l||}{} & \multicolumn{2}{c||}{Cooperative}        & \multicolumn{2}{c||}{Introduction of SI agent} & Re-adaptation \\
        \multicolumn{2}{l||}{Task} & \multicolumn{1}{l|}{w/ comms} & w/o comms & w/ adv comms  & w/o adv comms   & w/ adv comms  \\ \hline \hline
        \multirow{2}{*}{Coverage}& C & 67.1 $\pm$ 2.8 &63.2 $\pm$ 5.9 &45.6 $\pm$ 5.2 &58.6 $\pm$ 2.3 &60.8 $\pm$ 2.7 \\ 
& SI & N/A &N/A &103.7 $\pm$ 21.1 &45.5 $\pm$ 10.0 &32.9 $\pm$ 13.3 \\ \hline 
\multirow{2}{*}{\begin{tabular}[c]{@{}l@{}}Split\\Coverage\end{tabular}}& C & 52.9 $\pm$ 0.5 &38.9 $\pm$ 11.0 &34.8 $\pm$ 3.0 &47.5 $\pm$ 1.9 &50.8 $\pm$ 1.4 \\ 
& SI & N/A &N/A &90.9 $\pm$ 15.0 &27.6 $\pm$ 9.4 &10.4 $\pm$ 6.8 \\ \hline 
\multirow{2}{*}{\begin{tabular}[c]{@{}l@{}}Path\\planning\end{tabular}}& C & 29.1 $\pm$ 6.0 &22.4 $\pm$ 6.6 &8.9 $\pm$ 4.3 &28.0 $\pm$ 6.6 &27.1 $\pm$ 5.0 \\ 
    & SI & N/A &N/A &37.7 $\pm$ 10.1 &17.8 $\pm$ 19.8 &5.6 $\pm$ 13.8
    \end{tabular}
    \caption{Average return for all agents of each group (cooperative and self-interested (SI)) over 100 episodes at the end of training for all experiments given with a $1\sigma$ standard deviation. Rows show the tasks and agent groups, cooperative (C)  and self-interested (SI); columns show the three experiment types, (1)-(3), and communication variants, as described in~\autoref{sec:experiments}.
    \label{tab:results}}
\end{table}

\begin{figure}[htb]
    \centering
    \begin{subfigure}[t]{\textwidth}
        \centering
        \includegraphics[width=0.32\textwidth]{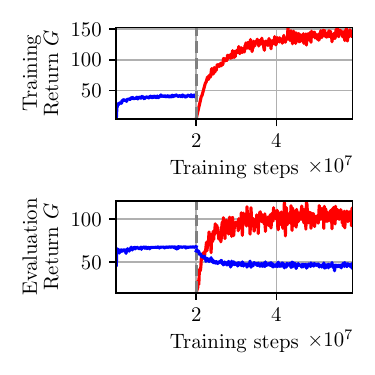}
        \includegraphics[width=0.32\textwidth]{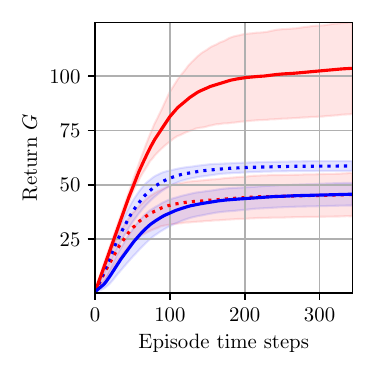}
        \includegraphics[width=0.32\textwidth]{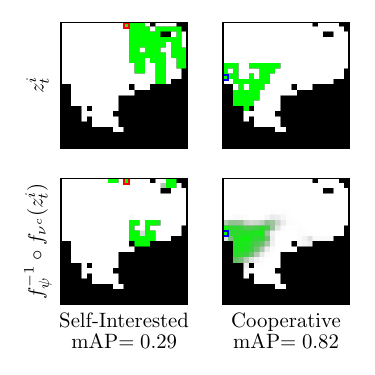}
        \caption{Coverage path planning in non-convex environment.}
        \label{fig:results_coverage}
    \end{subfigure}
    \hfill
    \begin{subfigure}[t]{\textwidth}
        \centering
        \includegraphics[width=0.32\textwidth]{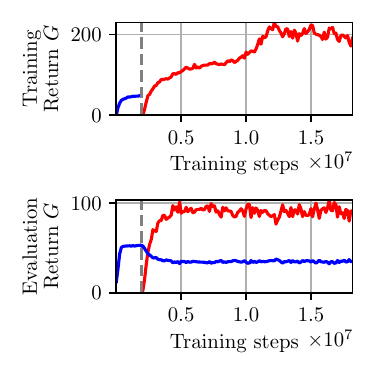}
        \includegraphics[width=0.32\textwidth]{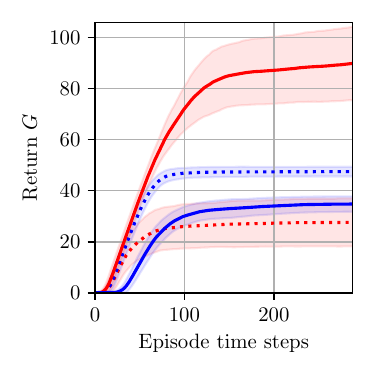}
        \includegraphics[width=0.32\textwidth]{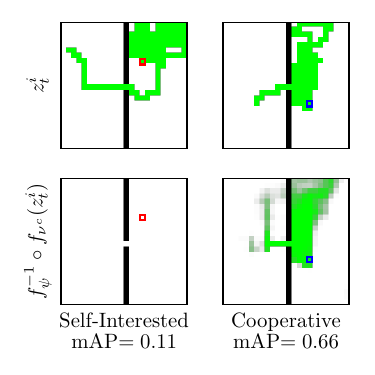}
        \caption{Coverage path planning in split environment.}
        \label{fig:results_split}
    \end{subfigure}
    \hfill
    \begin{subfigure}[t]{\textwidth}
        \centering
        \includegraphics[width=0.32\textwidth]{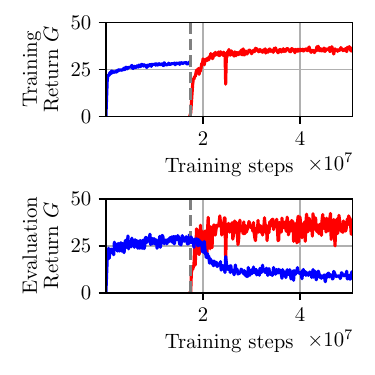}
        \includegraphics[width=0.32\textwidth]{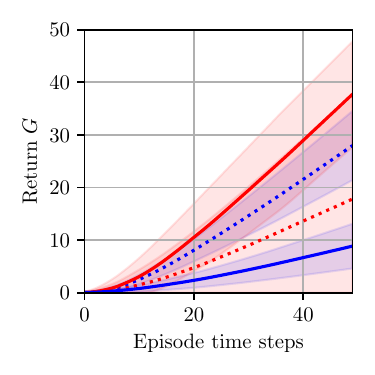}
        \includegraphics[width=0.32\textwidth]{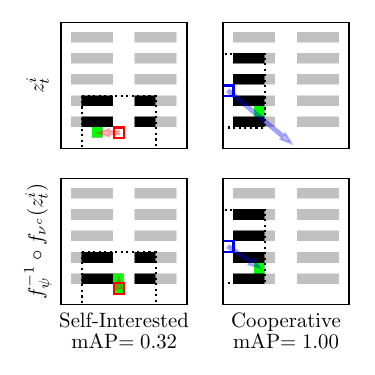}
        \caption{Path planning.}
        \label{fig:results_path}
    \end{subfigure}
    \caption{The sub-panels show results for the three considered tasks. \textit{1st column:} Mean episodic reward normalized per agent during training; blue for the cooperative agents and red for the self-interested agent. The top plot shows the sequential training of the cooperative team followed by the training of the self-interested agent while holding the cooperative policies fixed. The bottom plot shows the reward during training evaluation for a fixed episode length. \textit{2nd column}: Mean test reward per agent group (self-interested or cooperative) over 100 episodes, throughout an episode. The solid curves show the return with adversarial communications and the dashed curves without adversarial communications. All curves are given with a $1\sigma$ standard deviation. \textit{3rd column}: Visualization of the white-box analysis on an example. The top row shows the true local observation for the self-interested agent and for a single cooperative agent. The bottom row shows the reconstruction of the message according to the trained interpreter.}
    \label{fig:results}
\end{figure}

\textbf{Results.}
The results are summarized in~\autoref{tab:results} and visualized in~\autoref{fig:results}.
In a purely cooperative team (first two columns), agents are able to improve their average performance by utilizing explicit communication to coordinate. 
After the introduction of a self-interested agent, average performance for the cooperative team decreases; this loss is significant in the case where adversarial communication is enabled. The self-interested agent is able to significantly outperform the cooperative team.
When communicating, its performance improves by 128\% for non-convex coverage, by 229\% for split coverage, and by 112\% for path planning.
After re-adaptation, the cooperative team is able to recoup its performance loss, and reach 
a level that is on par with the purely cooperative case (with communication).
\autoref{fig:results} shows performance during training (first column) and final testing performance over an episode (second column). For all tasks, the \gls{mAP} of the white-box interpreter on the test set is significantly higher for the cooperative agents than for the self-interested agent, indicating that the self-interested agent learns an encoding that differs from the cooperative policy's encoding of local observations. We include additional experiments in \autoref{sec:sup:additional_experiments}.

\textbf{Discussion.}
Allowing the self-interested agent to learn its policy while holding all other agents' policies fixed leads to manipulative behavior that is made possible through adversarial communication. We showed that this observation is valid across different task settings, with performance improvements in the range of 112\%--229\% for the self-interested agent when communication is enabled.
Conversely, adversarial communication is neutralized when other agents are able to adapt to the self-interested policy.
Overall, we demonstrate that adversarial communication emerges when local rewards are drawn from a finite pool, or when resources are in contention; self-interested agents that communicate manipulatively, however, are not necessarily adversarial by design, they are simply programmed to disregard other agents' rewards.

\section{Conclusion}
We proposed a novel model for learning to communicate in multi-agent systems with multiple, potentially conflicting incentives that are driven by local agent-specific rewards. The main attribute of our model is that it is capable of accommodating multiple agent objectives while maintaining a common differentiable communication channel.
We demonstrated the emergence of adversarial communication and listed conditions under which this was observed. Post-hoc interpretations indicated devious encodings in messages sent through self-interested agents. 
Future work will address co-optimization schemes, the study of equilibria, and a generalization of the proposed methods to arbitrary proportions of cooperative vs. self-interested agents.

\clearpage

\acknowledgments{We gratefully acknowledge the support of ARL grant DCIST CRA W911NF-17-2-0181, and the Engineering and Physical Sciences Research Council (grant EP/S015493/1). The research is funded in part by Arm, and in part by Amazon.com Inc. Their support is gratefully acknowledged.}

\small{\bibliography{bibliography}}  

\begin{appendices}

\section{Proofs}
\label{sec:sup:proofs}

\begin{lemma}
Given an actor-critic algorithm with a compatible TD(1) critic that follows the cooperative policy gradient
\begin{align}
    g^i_k = \E_{\bbpi} \left[ \sum_{j\in\ccalV} \nabla_\theta \log \pi^j_{\theta}(a^j | z) A^i(s) \right]
\end{align}
for each agent $i \in \ccalV$ at each iteration $k$, this gradient converges to a local maximum of the expected sums of returns of all agents with probability one. 
\end{lemma}
\begin{myproofnoname}
The total gradient applied to $\theta$ is given by
\begin{align}
    g &= \E_{\bbpi} \left[ \sum_{i\in\ccalV} \sum_{j\in\ccalV} \nabla_\theta \log \pi^j_{\theta}(a^j | z) A^i(s) \right] = \E_{\bbpi} \left[ \sum_{i\in\ccalV} A^i(s) \sum_{j\in\ccalV} \nabla_\theta \log \pi^j_{\theta}(a^j | z) \right] \\
     &= \E_{\bbpi} \left[ \sum_{i\in\ccalV} A^i(s) \nabla_\theta \log \prod_{j\in\ccalV} \pi^j_{\theta}(a^j | z) \right] 
    = \E_{\bbpi} \left[ \sum_{i\in\ccalV} A^i(s) \nabla_\theta \log \bbpi_{\theta}(a | z) \right].
\end{align}
If we consider the sum of rewards $r_t = \sum_{i \in \ccalV} r_t^i$ as the joint reward obtained by the joint policy $\bbpi$, the joint advantage estimate is $A(s) = \sum_{i\in\ccalV}A^i(s)$.
Hence the individual policy gradients lead to a joint policy gradient, which is known to converge to a local maximum of the expected return $G_t = \sum_{i \in \ccalV} G^i_t$ if {\it(i)} $\bbpi$ is differentiable, {\it(ii)} the update timescales are sufficiently slow, and {\it(iii)} the advantage estimate uses a representation compatible with $\bbpi$~\cite{konda2000actorcritic}.
\end{myproofnoname}

\begin{lemma}
Given an actor-critic algorithm with a compatible TD(1) critic that follows the self-interested policy gradient
\begin{align}
    g^n_k = \E_{\bbpi} \left[ \sum_{j\in\ccalV} \nabla_{\theta^n} \log \pi^j_{\theta^c\theta^n}(a^j | z) A^n(s) \right]
\end{align}
for a self-interested agent $n \in \ccalV$ at each iteration $k$, this gradient converges to a local maximum of the expected returns of the self-interested agent with probability one. 
\end{lemma}
\begin{myproofnoname}
The gradient applied to $\theta$ is given by
\begin{align}
    g^n &= \E_{\bbpi} \left[ \sum_{j\in\ccalV} \nabla_{\theta^n} \log \pi^j_{\theta^c\theta^n}(a^j | z) A^n(s) \right]
    = \E_{\bbpi} \left[ A^n(s) \sum_{j\in\ccalV} \nabla_{\theta^n} \log \pi^j_{\theta^c\theta^n}(a^j | z)  \right] \\
    &= \E_{\bbpi} \left[ A^n(s) \nabla_{\theta^n} \log \prod_{j\in\ccalV}  \pi^j_{\theta^c\theta^n}(a^j | z)  \right]
    = \E_{\bbpi} \left[ A^n(s) \nabla_{\theta^n} \log \bbpi_{\theta^c\theta^n}(a | z) \right].
\end{align}
If we consider the self-interested agent's reward $r_t = r_t^n$ as the joint reward obtained by the joint policy $\bbpi$, the joint advantage estimate is $A(s) = A^n(s)$.
Hence, the self-interested policy gradient leads to a joint policy gradient which is known to converge to a local maximum of the expected return $G_t = G^n_t$ if {\it(i)} $\bbpi$ is differentiable, {\it(ii)} the update timescales are sufficiently slow, and {\it(iii)} the advantage estimate uses a representation compatible with $\bbpi$~\cite{konda2000actorcritic}.
\end{myproofnoname}

\section{Learning Algorithms}
\label{sec:sup:learning_algorithms}
\autoref{alg:coop} and ~\autoref{alg:adv} detail the algorithms for the Cooperative Policy Gradient and Self-Interested Policy Gradient, respectively.

\begin{algorithm}[htb]
\SetAlgoLined
  \textbf{Input:} Initial policy parameters $\theta$ and value parameters used to estimate the advantage \\
  \For{$k \leftarrow 1, 2, \dots$}{
    Collect set of trajectories $\ccalD_t$ by running policies $\pi_\theta^i$ \\
    Compute advantage estimates $\hat{A}^i(s_t)$ for all time steps \\
    Estimate the policy gradient as $g_k = \sum_{i \in \ccalV} g_k^i$ with $g_k^i = \frac{1}{|\ccalD_k|} \sum_{\tau \in \ccalD_k} \sum_{t \in \tau} \sum_{j\in\ccalV} \nabla_\theta \log \pi^j_{\theta}(a_t^j | z_t) A^i(s_t)$ \\
    Compute policy update $\theta \gets \theta + \alpha_k g_k$ \\
    Fit value functions \\
  }
  \caption{Cooperative Policy Gradient}
  \label{alg:coop}
\end{algorithm}

\begin{algorithm}[htb]
\SetAlgoLined
  \textbf{Input:} Initial policy parameters $\theta^n$ and value parameters used to estimate the advantage \\
  \For{$k \leftarrow 1, 2, \dots$}{
    Collect set of trajectories $\ccalD_t$ by running policies $\pi_{\theta^c\theta^n}^i$ \\
    Compute advantage estimate $\hat{A}^n(s_t)$ for all time steps \\
    Estimate the policy gradient as $g_k^n = \frac{1}{|\ccalD_k|} \sum_{\tau \in \ccalD_k} \sum_{t \in \tau} \sum_{j\in\ccalV} \nabla_{\theta^n} \log \pi^j_{\theta^c\theta^n}(a_t^j | z_t) A^n(s_t)$ \\
    Compute policy update $\theta^n \gets \theta^n + \alpha_k g^n_k$ \\
    Fit value function \\
  }
  \caption{Self-Interested Policy Gradient}
  \label{alg:adv}
\end{algorithm}

\section{Additional Experiments}
\label{sec:sup:additional_experiments}
In this section, we provide additional results. In particular, we show \textit{(i)} how cooperative agents can re-adapt to counteract adversarial communication (i.e., \textit{``fool me once, shame on you; fool me twice, shame on me''}), \textit{(ii)} the nature of communicated messages and their effect on agents' internal representations, and \textit{(iii)} that adversarial communication only arises when there is resource contention.

\begin{figure}[tb]
    \centering
    \begin{subfigure}[t]{\textwidth}
        \centering
        \includegraphics[width=0.49\textwidth]{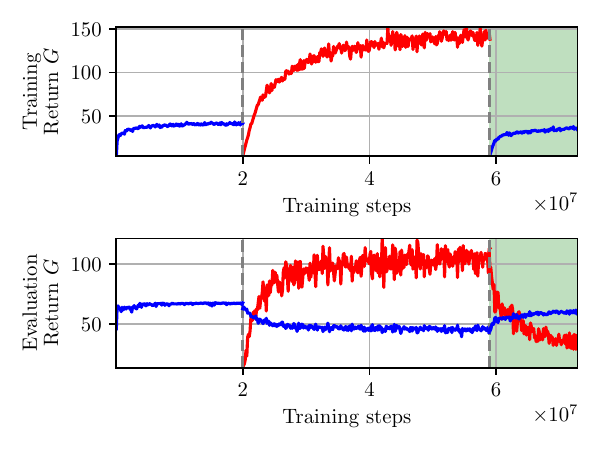}
        \includegraphics[width=0.49\textwidth]{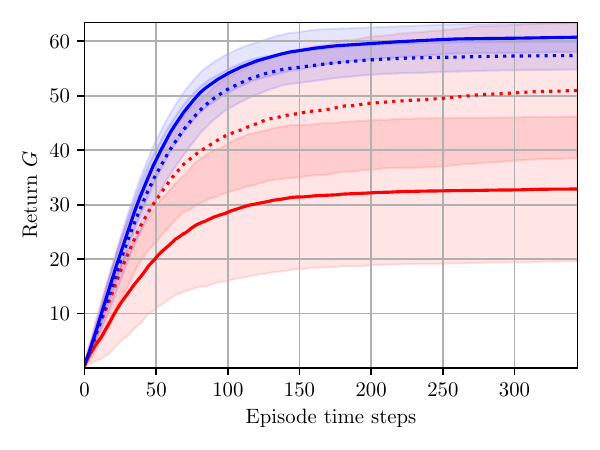}
        \caption{Coverage non-convex environment.}
        \label{fig:results_coverage_detailed}
    \end{subfigure}
    \hfill
    \begin{subfigure}[t]{\textwidth}
        \centering
        \includegraphics[width=0.49\textwidth]{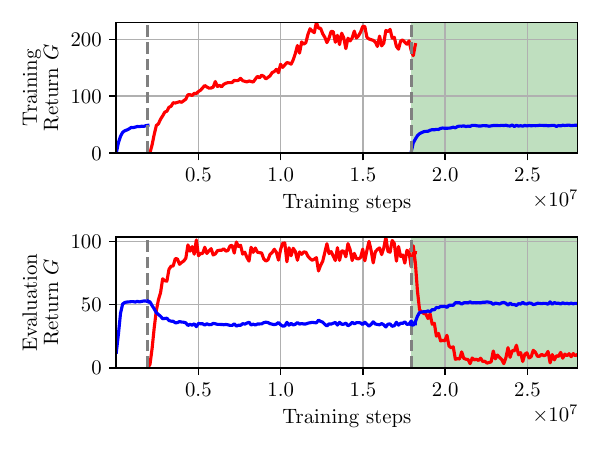}
        \includegraphics[width=0.49\textwidth]{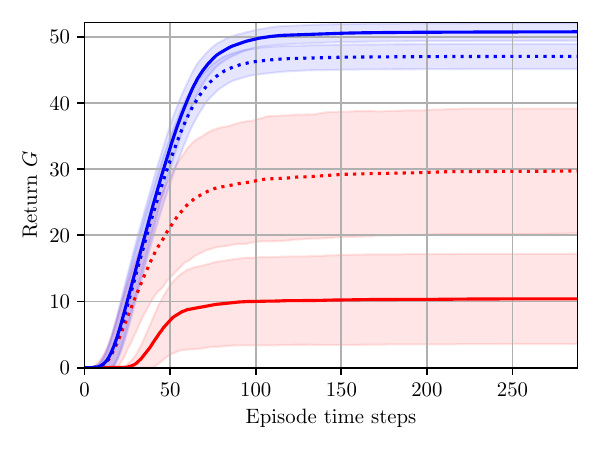}
        \caption{Coverage in split environment.}
        \label{fig:results_split_detailed}
    \end{subfigure}
    \hfill
    \begin{subfigure}[t]{\textwidth}
        \centering
        \includegraphics[width=0.49\textwidth]{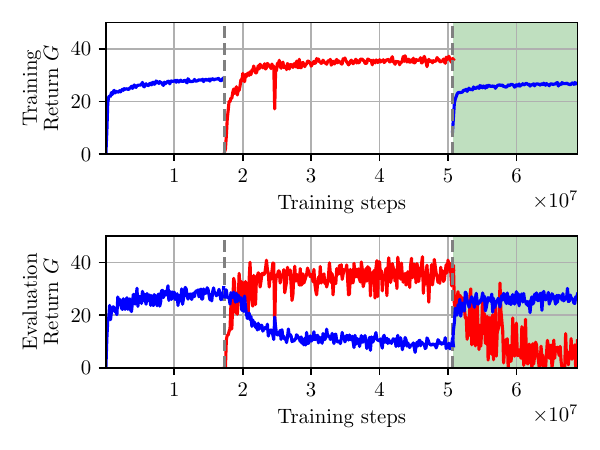}
        \includegraphics[width=0.49\textwidth]{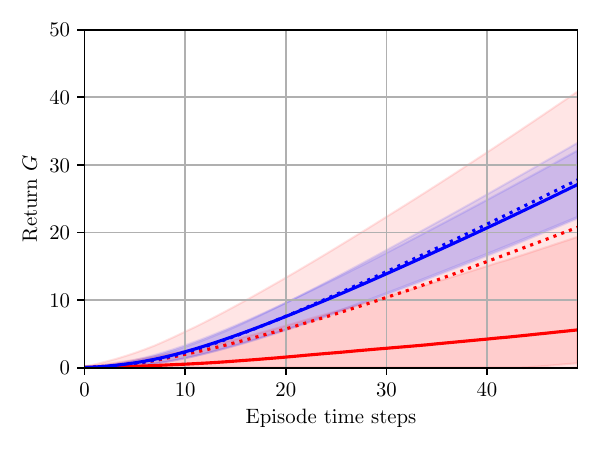}
        \caption{Path planning.}
        \label{fig:results_path_detailed}
    \end{subfigure}
    \caption{Cooperative agents continue their training, and learn to counteract adversarial communications, for all tasks in (a)-(c). \textit{Left:} Performance during training and evaluation. Re-adaptation occurs during the \textit{green} phase. \textit{Right:} Performance during testing with the final model throughout an episode. We perform 100 episode runs; the solid curves show the return with adversarial communications and the dashed curves without adversarial communications. All curves are given with a 1$\sigma$ standard deviation.}
    \label{fig:results_readaptation}
\end{figure}

\autoref{fig:results_readaptation} demonstrates how cooperative agents learn to counteract adversarial communications, when allowed to continue their training in the presence of the self-interested agent. The test results show no performance loss for the cooperative agents when adversarial communication is enabled (solid blue curve), whereas the self-interested agent now performs significantly worse.

\begin{figure}[tb]
    \centering
    \includegraphics[width=\textwidth]{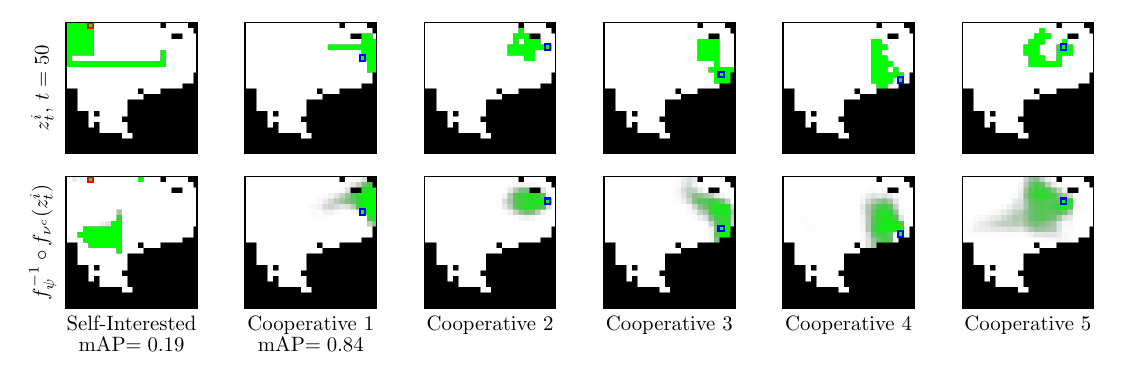}
    \includegraphics[width=\textwidth]{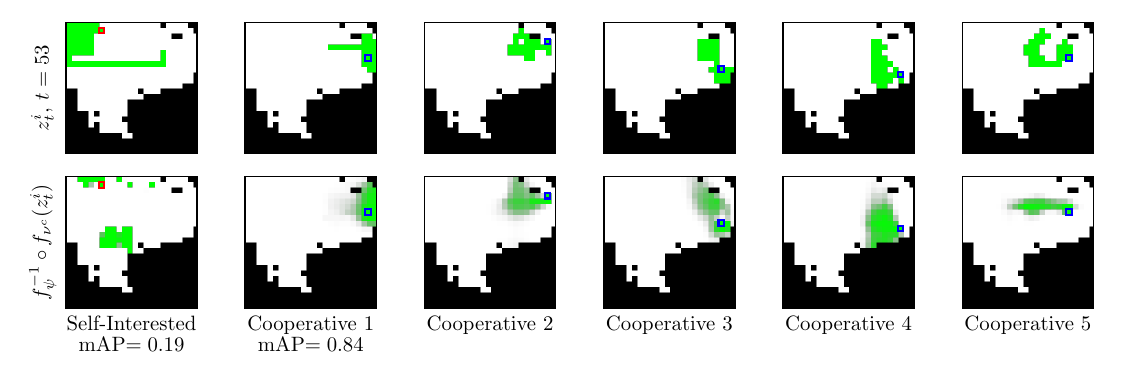}
    \includegraphics[width=\textwidth]{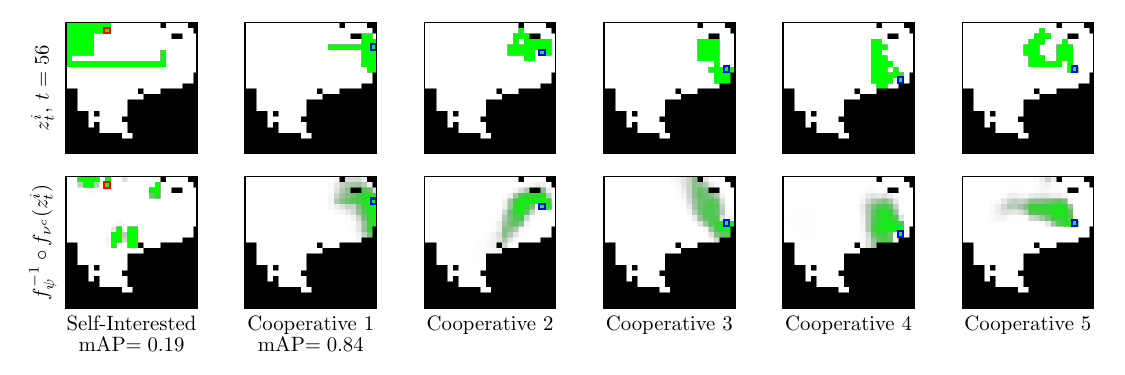}
    \includegraphics[width=\textwidth]{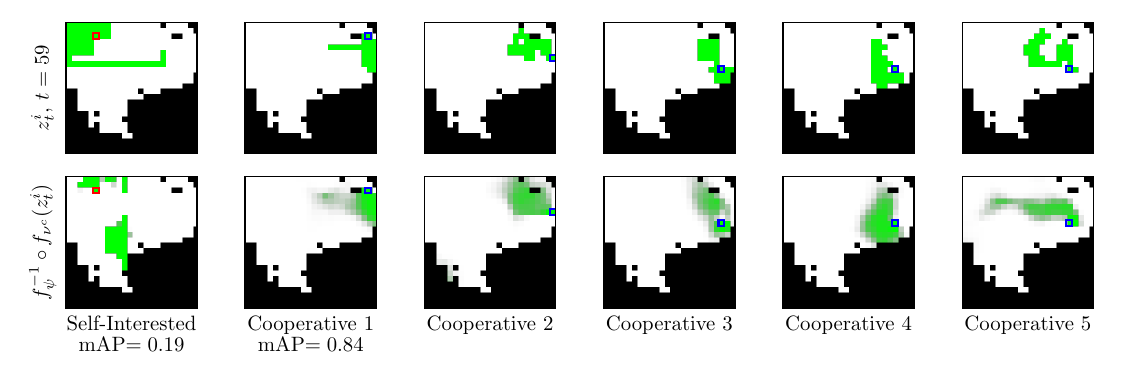}
    \caption{Sequence of interpreted messages for non-convex coverage (with adversarial communication). Column 1 shows the self-interested agent (in red) and columns 2-6 show five cooperative agents (in blue). We consider 4 successive instances in time (top to bottom). For each time instance, we show two rows: the top row shows the true local coverage, and the bottom row shows the communicated local coverage. The adversarial agent clearly sends false information, whereas cooperative agents are truthful.
    \label{fig:results_interpreter_sequence_coverage}}
\end{figure}

\begin{figure}[tb]
    \centering
    \includegraphics[width=\textwidth]{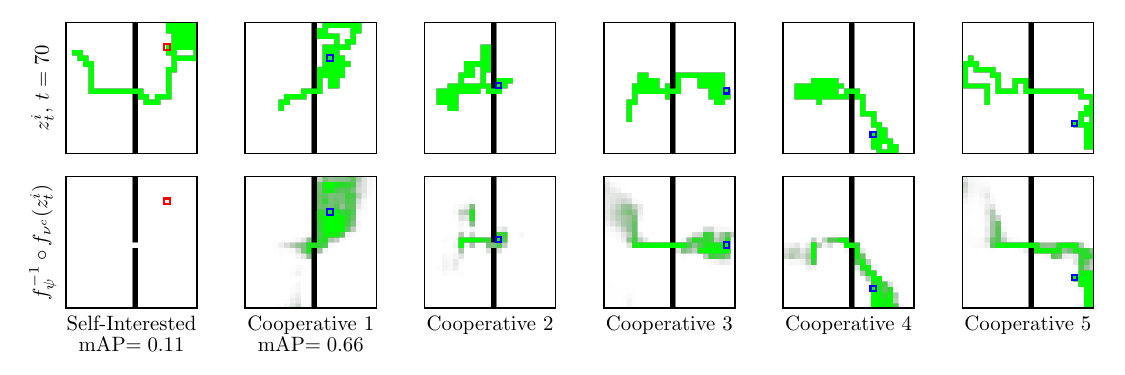}
    \includegraphics[width=\textwidth]{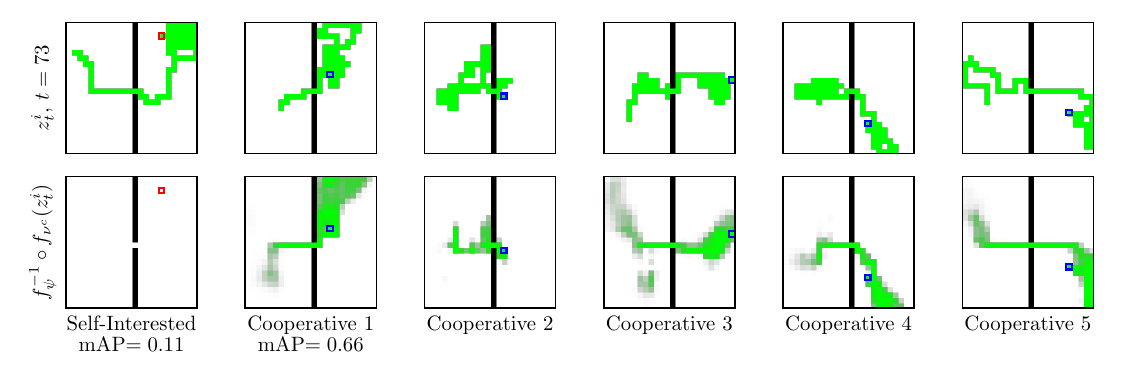}
    \includegraphics[width=\textwidth]{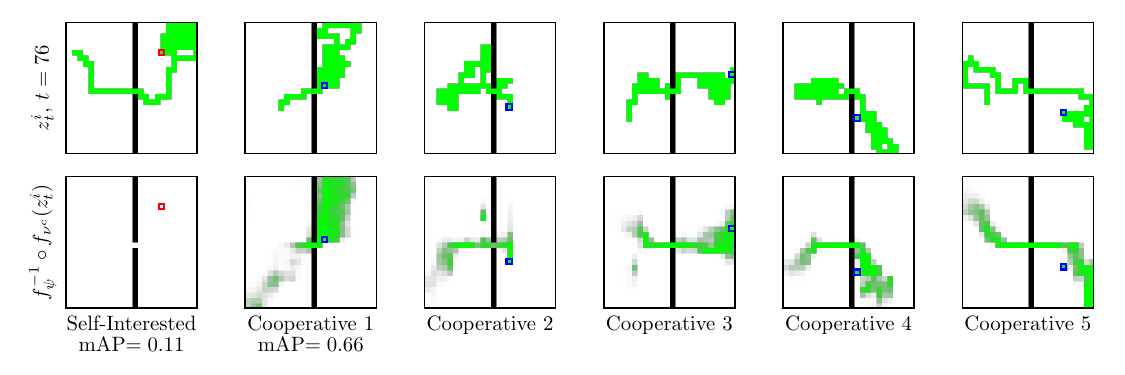}
    \includegraphics[width=\textwidth]{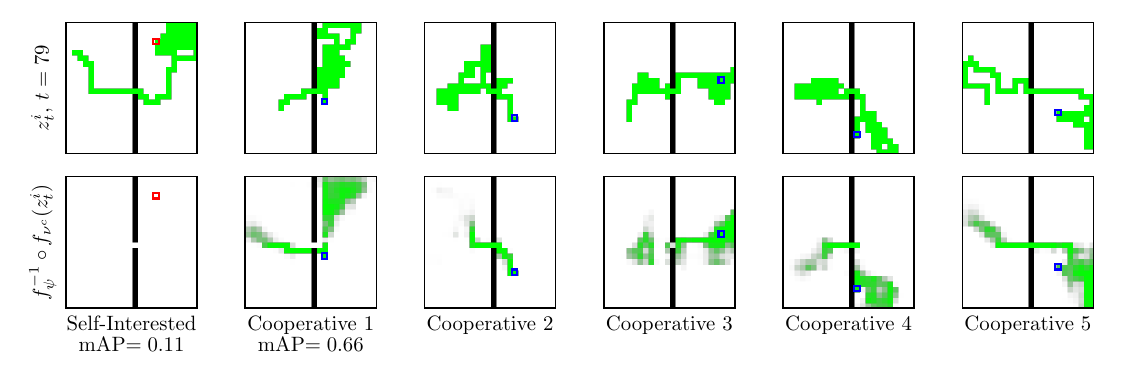}
    \caption{Sequence of interpreted messages for split coverage (with adversarial communication). Column 1 shows the self-interested agent (in red) and columns 2-6 show five cooperative agents (in blue). We consider 4 successive instances in time (top to bottom). For each time instance, we show two rows: the top row shows the true local coverage, and the bottom row shows the communicated local coverage. The adversarial agent clearly sends false information, whereas cooperative agents are truthful.}
    \label{fig:results_interpreter_sequence_split}
\end{figure}

\begin{figure}[tb]
    \centering
    \includegraphics[width=\textwidth]{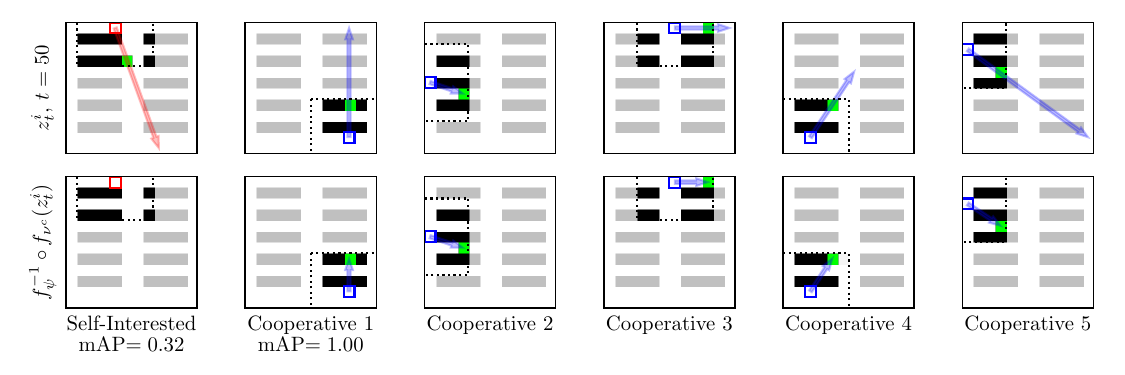}
    \includegraphics[width=\textwidth]{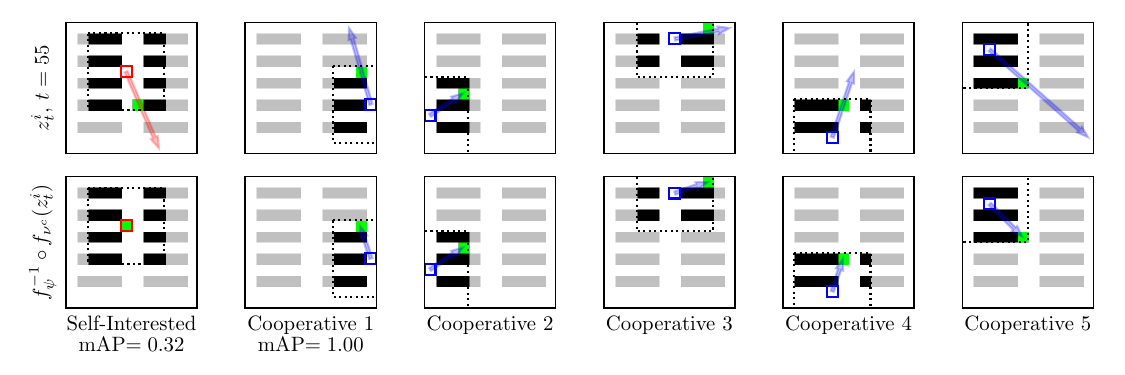}
    \includegraphics[width=\textwidth]{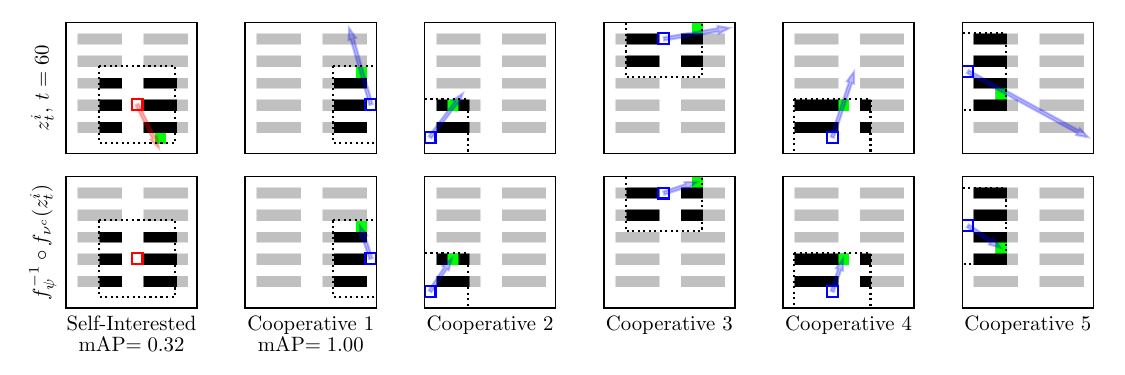}
    \includegraphics[width=\textwidth]{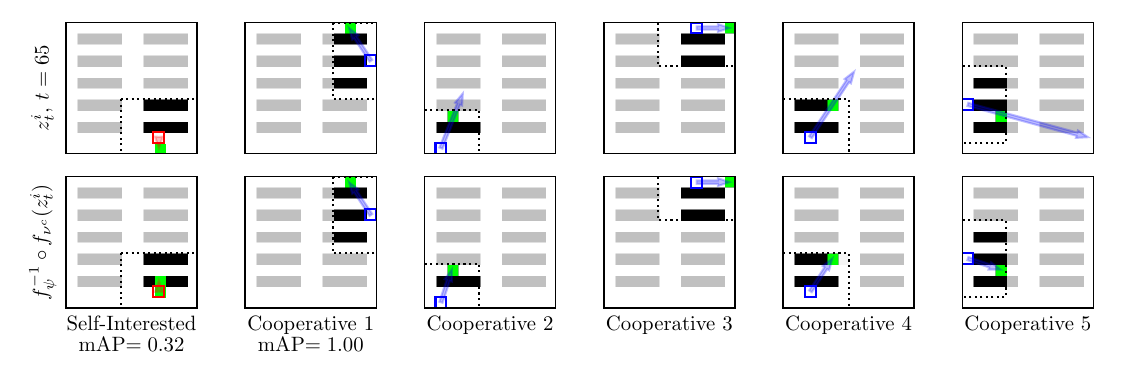}
    \caption{Sequence of interpreted messages for the path planning task (with adversarial communication). Column 1 shows the self-interested agent (in red) and columns 2-6 show five cooperative agents (in blue). We consider 4 successive instances in time (top to bottom). For each time instance, we show two rows: the top row shows the agent's true goal, and the bottom row shows the communicated information. When the goal lies outside the field-of-view, it is projected to its perimeter. The adversarial agent clearly sends false information about its goal, whereas cooperative agents are truthful.}
    \label{fig:results_interpreter_sequence_path}
\end{figure}

\autoref{fig:results_interpreter_sequence_coverage},~\autoref{fig:results_interpreter_sequence_split} and~\autoref{fig:results_interpreter_sequence_path} show the interpretation of messages sent, for the three considered tasks, respectively.
Overall, the panels confirm that messages sent by the cooperative agents are truthful, yet that messages sent by the self-interested agent are false.

\begin{figure}[tb]
    \centering
    \includegraphics[width=\textwidth]{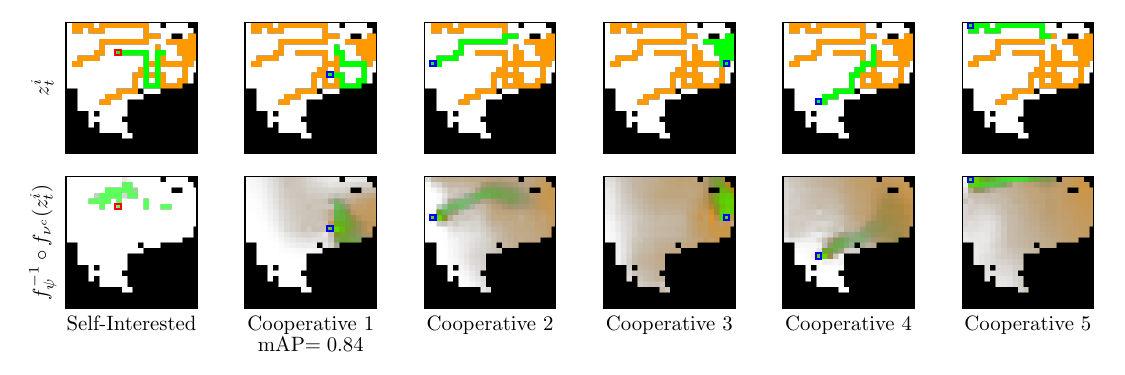}
    \includegraphics[width=\textwidth]{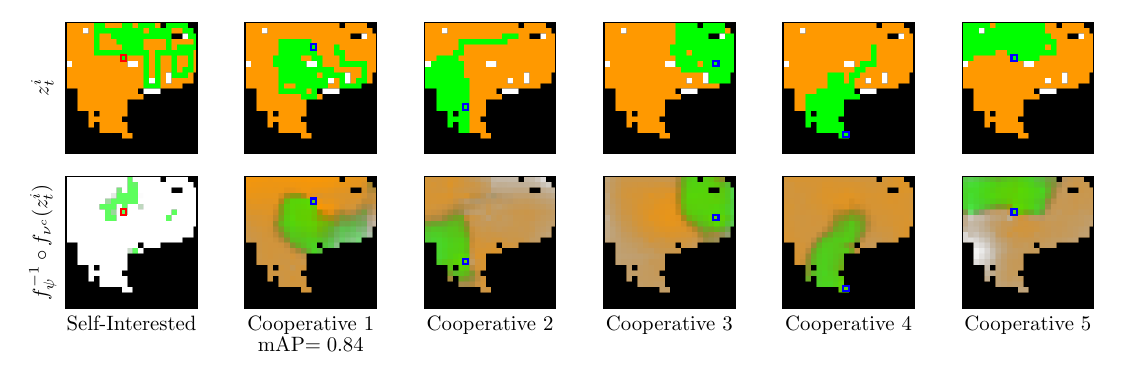}
    \includegraphics[width=\textwidth]{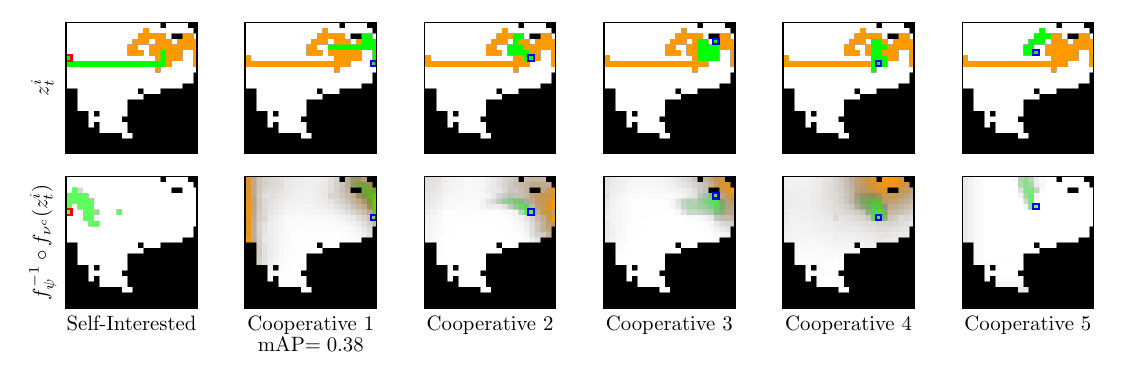}
    \includegraphics[width=\textwidth]{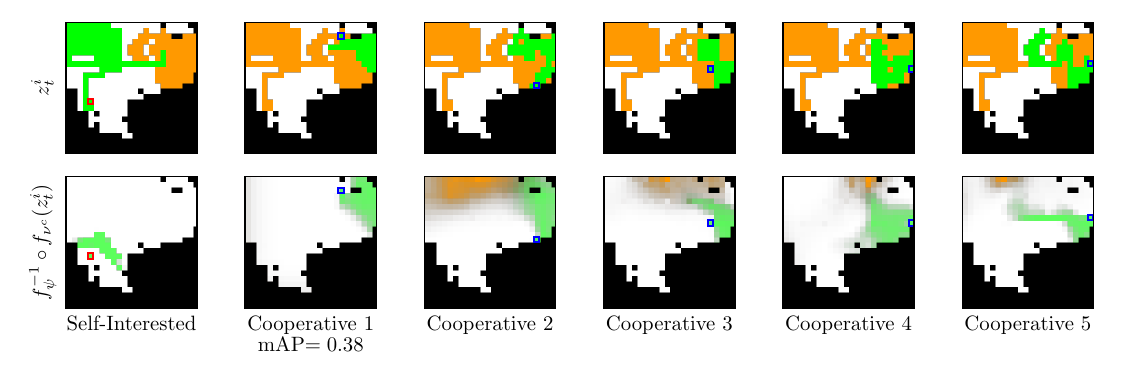}
    \caption{We train the interpreter on the output of the AGNN to reconstruct the local representation of the global coverage (in orange). Each agent's local coverage is shown in green. The first two result sets are obtained without adversarial communication, and the bottom two with. Column 1 shows the self-interested agent (in red) and columns 2-6 show five cooperative agents (in blue). The panels show how the self-interested agent is able to manipulate the local estimate of global coverage.}
    \label{fig:results_interpreter_sequence_global_cov}
\end{figure}

\begin{figure}[tb]
    \centering
    \includegraphics[width=\textwidth]{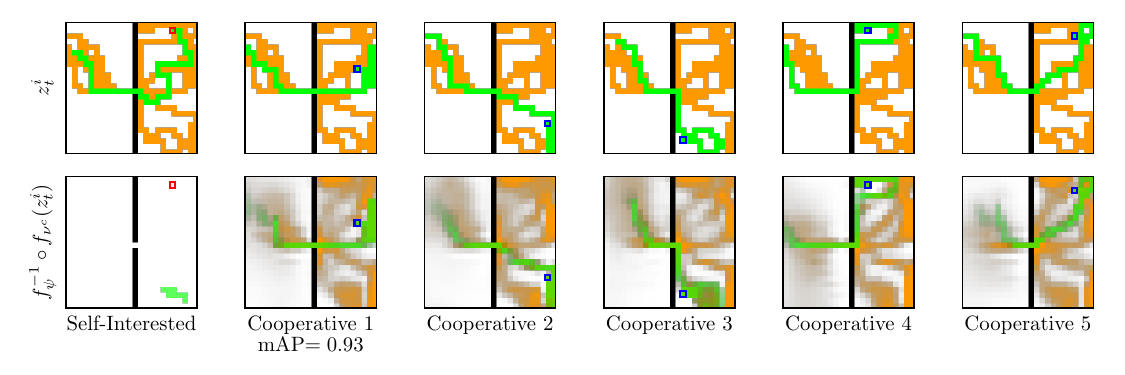}
    \includegraphics[width=\textwidth]{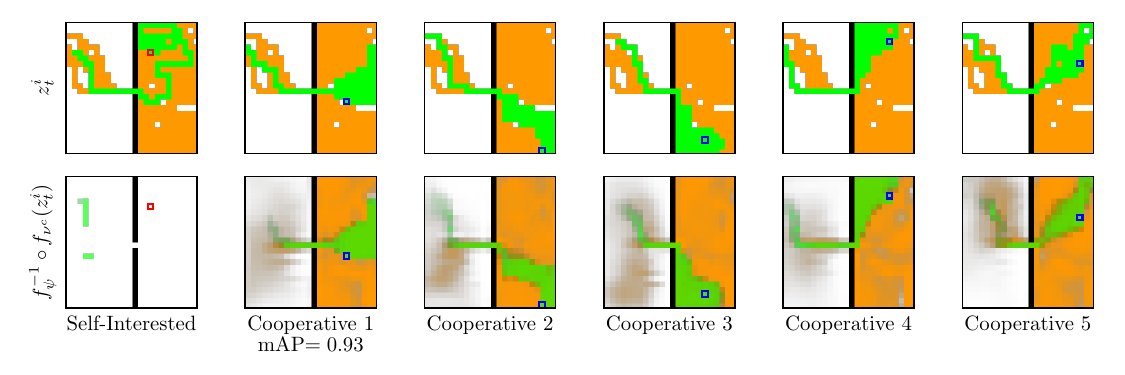}
    \includegraphics[width=\textwidth]{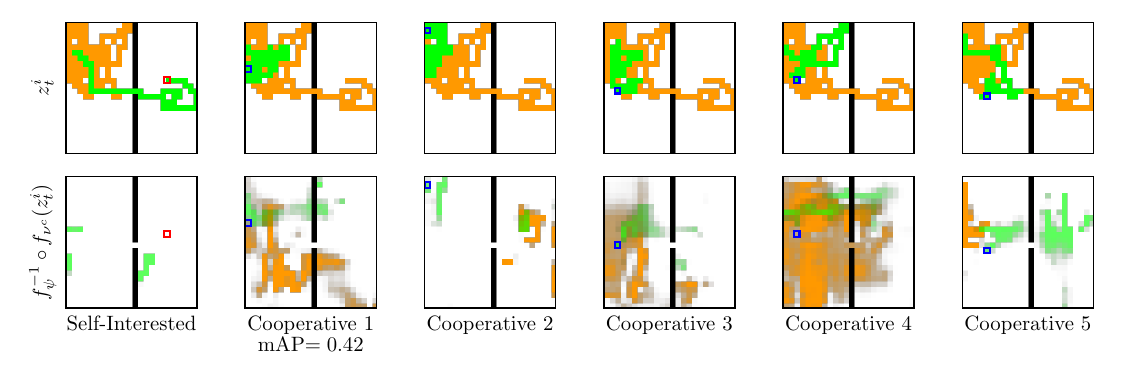}
    \includegraphics[width=\textwidth]{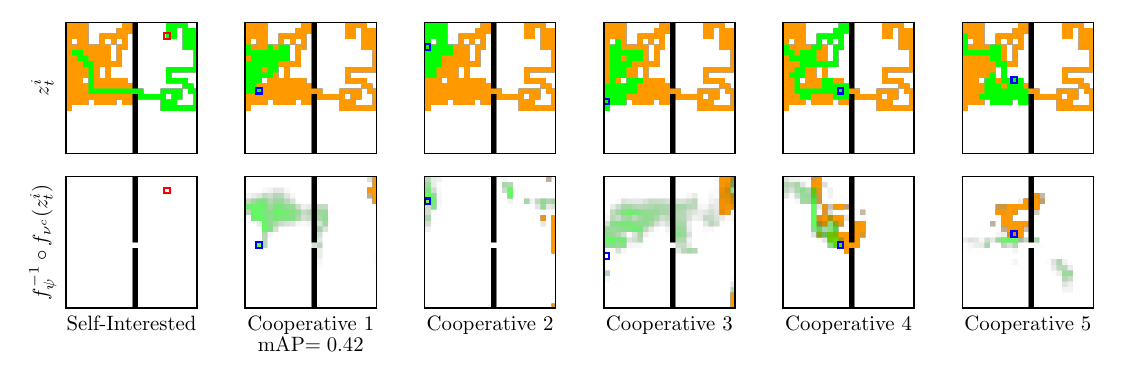}
    \caption{We train the interpreter on the output of the AGNN to reconstruct the local representation of the global coverage (in orange). Each agent's local coverage is shown in green. The first two result sets are obtained without adversarial communication, and the bottom two with. Column 1 shows the self-interested agent (in red) and columns 2-6 show five cooperative agents (in blue). The panels show how the self-interested agent is able to manipulate the local estimate of global coverage.}
    \label{fig:results_interpreter_sequence_global_split}
\end{figure}

\autoref{fig:results_interpreter_sequence_global_cov} and~\autoref{fig:results_interpreter_sequence_global_split} show how the messages sent by the self-interested agent impact the local estimates of global coverage. They demonstrate that when adversarial communication is enabled, agents' local estimates are manipulated to represent faulty information.

\begin{figure}[tb]
    \centering
    \begin{minipage}{.5\textwidth}
        \centering
        \raisebox{0.1\height}{\includegraphics[width=0.6\textwidth]{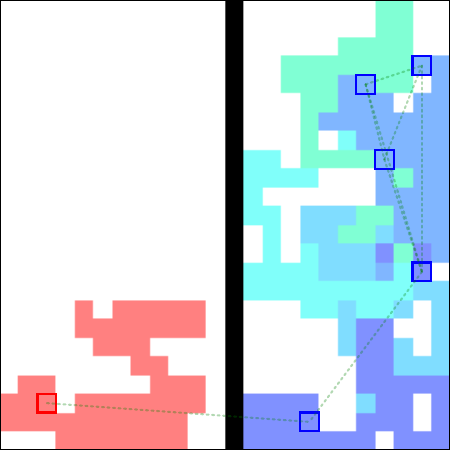}}
    \end{minipage}%
    \begin{minipage}{0.5\textwidth}
        \centering
        \includegraphics[width=\textwidth]{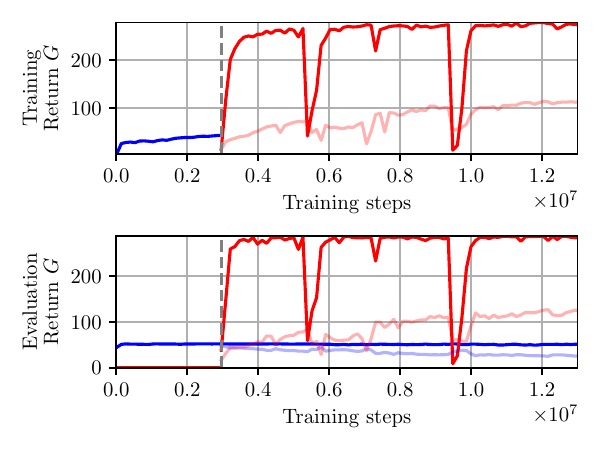}
    \end{minipage}
    \begin{minipage}{.5\textwidth}
        \centering
    \includegraphics[width=\textwidth]{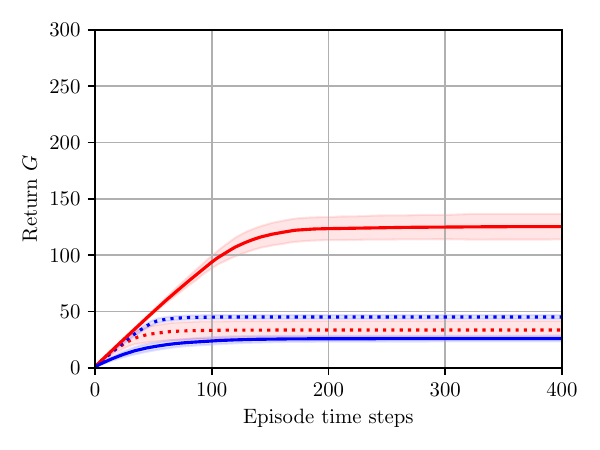}
    \end{minipage}%
    \begin{minipage}{0.5\textwidth}
        \centering
    \includegraphics[width=\textwidth]{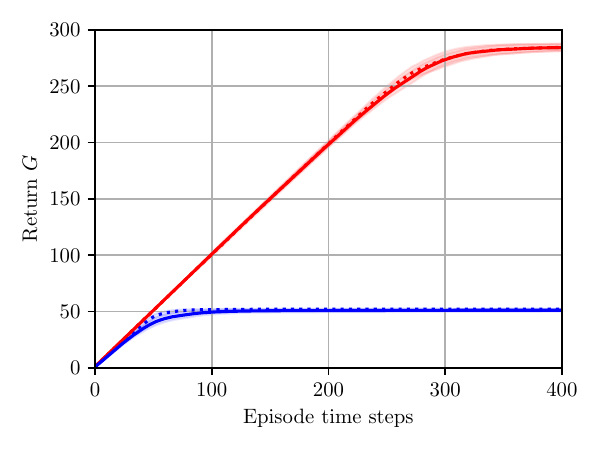}
    \end{minipage}
    \caption{Adversarial training in a non-competitive environment where resources are non-shared. Red curves show the return for the self-interested agent and blue curves for the cooperative agent group. \textit{Top left:} An example of the environment, which is physically split (the communication graph remains connected). \textit{Top right:} The solid curves show the training and evaluation return of the non-competitive environment (self-interested agent placed in left half) and the light curves show the return for the competitive environment (self-interested agent placed together with cooperative agents in the right half). The plot is separated in two areas, left shows cooperative training and right self-interested training. The blue solid curve in the evaluation plot continues at the same level over the course of the training while the red curve quickly rises, indicating that no adversarial communication is learned as the cooperative agents consistently cover the same area on the right side while the self-interested agent learns to cover the left side on its own. In contrast, the blue light curve decreases while the red light curve increases, indicating that the self-interested agent learns adversarial communications as it gradually covers more area that, in turn, cannot be covered by the cooperative agents. \textit{Bottom:} Both plots in the bottom show the testing performance throughout 100 episode runs on the final model. We provide $1\sigma$ error bars; the solid curves show the return with adversarial communications and the dashed curves without adversarial communications. \textit{Bottom left:} Testing performance for baseline in competitive agent placement (self-interested agent shares right half with cooperative agents). There is a significant difference in performance with and without adversarial communications. \textit{Bottom right:} When placing the self-interested agent in the separated half as depicted in top left, the difference in performance with and without adversarial communications disappears. Note that the evaluation magnitude of the return of the cooperative agents and self-interested agent differs because the area is split in half while the team sizes are unbalanced, resulting in a differing per-agent average return.}
    \label{fig:results_coverage_no_competition}
\end{figure}

\autoref{fig:results_coverage_no_competition} shows an example of a physically split, non-shared environment. The results demonstrate that in environments without resource contention, there is no performance difference \textit{with} and \textit{without} adversarial communication. For this figure, we trained an interpreter on the final output of the \Gls{AGNN}.

\clearpage 
\section{Implementation Details}
\label{sec:sup:impl_details}
\paragraph{Hyperparameters.}
\begin{table}[tb]
\begin{tabular}{c|c|c|c|c|c|c|c}
 & \begin{tabular}[c]{@{}c@{}}Learning\\ Rate\end{tabular} & $\gamma$ & $\epsilon$ & $\lambda$ & \begin{tabular}[c]{@{}c@{}}Training\\ Batch Size\end{tabular} & \begin{tabular}[c]{@{}c@{}}Minibatch\\ Size\end{tabular} & \begin{tabular}[c]{@{}c@{}}SGD\\ Iterations\end{tabular} \\ \hline
Coverage (both) & $5 \cdot 10^{-4}$ & $0.9$ & $0.2$ & $0.95$ & $5000$ & $1000$ & $5$ \\ \hline
Path planning & $4 \cdot 10^{-4}$ & $0.99$ & $0.2$ & $0.95$ & $5000$ & $1000$ & $5$
\end{tabular}
\vspace{0.5cm}
\caption{Overview of training and PPO hyperparameters. The discount factor is denoted as $\gamma$, the PPO clipping parameter is denoted as $\epsilon$ and the GAE bias-variance parameter is denoted as $\lambda$. SGD iterations refer to the number of consecutive stochastic gradient descent optimization iterations for each training batch.}
\label{tab:hyperparameters}
\end{table}

We adapt Proximal Policy Optimization (PPO) to integrate our policy gradients and optimize using minibatch stochastic gradient descent as implemented in RLlib \cite{liang_2018}. The chosen hyperparameters for both experiments are shown in \autoref{tab:hyperparameters}.

\paragraph{Environment setup.} 

The environment of size $W \in \naturals$, $H \in \naturals$ is populated with $N$ agents $\ccalV=\{1, \dots, N\}$. Each agent $i$ is described at discrete time $t$ by its position $\bbp_t^i \in \naturals^{2}$ and a map of the environment $\bbM_t^i \in \booleans^{W \times H}$. 
Agents $i$ and $j$ can communicate to each other only if $\|\bbp_t^i-\bbp_t^i\| < d$, resulting in the communication graph $\ccalG_t$.
The graph $\ccalG_t$ is represented as adjacency matrix $\bbS_t$. The communication range $d$ is a hyperparameter of the environment. 
The adjacency matrix $\bbS_t$ is constructed for the adapted communication range and normalized to avoid exploding or vanishing gradients and therefore, numerical instability \cite{kipf_2017, gama_2019}.

At any time step, an agent can either move in one of four directions or wait. Agents' actions are constrained by the environment to prohibit collisions with obstacles or the world's margin. 
Each agent $i$ has a field of view (FOV) of width $W_{\fov} \in \naturals$ and height $H_{\fov} \in \naturals$. Each agent's partial observation can be described as tensor $z_t^i \in \reals^{2 \times W_{FOV} \times H_{FOV}}$ and consists of channels for a local world map $\bbM_t^i$ (i.e., obstacle positions) and, depending on the experiment, an optional map $\bbC_t^i$ containing the agent's local information such as coverage or goal. We integrate the agent's position $\bbp_t^i$ implicitly into the observation by shifting and cropping $\bbM_t^i$ and $\bbC_t^i$ to the agent-relative field of view so that the agent is centered. We pad the observation so that the map outside the dimensions of $\bbM_t^i$ is perceived as occupied with obstacles and the optional map $\bbC_t^i$ with zeros (i.e., no coverage and no goals).

\paragraph{Non-convex coverage.}
The coverage path planning problem is related to the \textit{covering salesman problem} in which an agent is required to visit all points in the target environment (an area or volume) as quickly as possible while avoiding obstacles \cite{galceran_2013}. Each agent keeps track of its local coverage $\bbC_t^i \in \booleans^{W \times H}$. The global coverage $\bbC_t \in \booleans^{W \times H}$ is the logical disjunction $\bbC_t = \overset{N}{\underset{i=1}{\vee}} \bbC_t^i$ of all agents' coverages.

The grid-world has a size of $W=H=24$ and is occupied with $40\%$ obstacles that are randomly generated while assuring connectivity of the environment.
Each agent receives a reward of 1 if it moves towards a cell that has not yet been covered by any agent. To incentivize fast coverage, in which all agents continuously cover new cells, and to increase sample efficiency, we use a dynamic episode length in which we terminate an episode if any agent has not covered a new cell for $10$ time steps.

We train the cooperative policy for 20M training environment time steps and the adversarial policy for 40M time steps. After completing the training, we evaluate with a fixed episode length $T = \lceil W \cdot H \cdot 0.6 \rceil = 346$, which is the minimum time required to visit every free cell for a single agent. 

\paragraph{Split coverage.}
The problem is similar to the non-convex coverage, but in this experiment, the agents only receive a reward for covering the \textit{right-hand side} of the environment. Hence, the agents have to learn to first move to that area and then coordinate to cover it.

The grid-world has a size of $W=H=24$. The environment has a fixed layout split vertically in two halves by a line of obstacles. The two halves are connected through a single cell. We initialize the agent's positions similarly to the previous experiment but constrain them to be placed in the left half.
The reward is similar to the non-convex coverage, but the agents are only rewarded for covering the right side of the environment. The early termination is similar, but is only triggered if at least one agent has reached the right side. Otherwise, the episode ends after a fixed time.

We train the cooperative policy for 2M training environment time steps and the adversarial policy for 15M time steps. We perform an evaluation with a fixed episode length $T = \lceil (W \cdot H)/2 \rceil = 288$ which is the time required to cover every cell in one half for a single agent. 

\paragraph{Path planning.}
In the path planning scenario, each agent is required to navigate to a labelled goal $\bbg_t^i \in \naturals^{2}$. In contrast to the coverage scenarios, only one agent can occupy a single cell at the same time. Since each agent is only aware of the relative environment layout and its own goal position, but not of other agent's positions, the agents have to communicate to coordinate and determine the most efficient set of paths.

The grid-world has a size of $W=H=12$. The environment resembles a warehouse layout with obstacles aligned in a regular grid. The agents are placed uniformly at random locations in the environment.
Each agent receives a local reward of 1 at each time step $t$ if $\bbp_t^i = \bbg_t^i$. Each episode has a horizon of $T=50$ time steps. Therefore the maximal possible mean reward per agent is $T$ if each agent immediately reaches its goal position.

We train the cooperative policy for 16M training environment time steps and the adversarial policy for 35M time steps. The episode length of $T$ during evaluation is similar to during training.

\paragraph{White-Box Analysis.} 
We collect $50$K training, $10$K validation and $5$K testing samples. To reduce correlation between samples, we sample at every time step $t$ with a probability of $10\%$ and discard all other samples.

The neural network architecture $f^{-1}_\psi$ consists of multiple combinations of 2D convolutions, LeakyReLU activation, 2-upsampling and zero padding where the specific hyperparameters depend on the input feature size and the desired output size. The final convolution has one output channel and a sigmoid activation. We use a binary cross-entropy loss. We are only interested in the features that are essential for an efficient collaboration between agents and therefore mask the loss and evaluation metrics correspondingly.

We train with a batch size of 64 for 200 training epochs. Every 10 training epochs, we evaluate the classification performance as \gls{mAP} on the validation set. We avoid overfitting by check-pointing the model with the best \gls{mAP} computed on the validation set. All interpreter visualizations in this paper are created from a subset of the testing samples.

\end{appendices}

\end{document}